\documentclass[twoside,11pt]{article}
\usepackage{booktabs} 
\usepackage{blindtext}
\usepackage{amsmath}
\usepackage{float}
\usepackage[table]{xcolor}

\usepackage[preprint]{jmlr2e}

\newcommand{\R}[1]{\mathbb{R}^{#1}}
\newcommand{\suchthat}{\ : \ }

\newcommand{\E}[1]{\mathbb{E}^{#1}}
\renewcommand{\H}[1]{\mathbb{H}^{#1}}
\renewcommand{\S}[1]{\mathbb{S}^{#1}}
\renewcommand{\P}{\mathcal{P}}
\newcommand{\M}{\mathcal{M}}

\newcommand{\X}{\mathbf{X}}
\newcommand{\x}{\mathbf{x}}
\newcommand{\y}{\mathbf{y}}
\renewcommand{\u}{\mathbf{u}}
\renewcommand{\v}{\mathbf{v}}
\newcommand{\p}{\mathbf{p}}
\newcommand{\W}{\mathbf{W}}
\newcommand{\w}{\mathbf{w}}
\newcommand{\D}{\mathbf{D}}

\renewcommand{\dot}[2]{\langle #1, #2 \rangle}

\newcommand{\PT}[3]{\operatorname{PT}_{#1\to#2}(#3)}
\NewDocumentCommand{\T}{o m}{%
  T_{#1}\IfValueTF{#2}{#2}{\mathcal{M}}%
}
\newcommand{\expmap}[2]{\exp_{#1}\left(#2\right)}
\newcommand{\logmap}[2]{\log_{#1}\left(#2\right)}
\newcommand{\origin}{\mathbf{\mu_0}}

\newcommand{\st}[2]{\mathfrak{st}^{#1}_{#2}}

\newcommand{\N}{\mathcal{N}}
\newcommand{\WN}{\mathcal{WN}}
\newcommand{\boldmu}{\mathbf{\mu}}
\newcommand{\boldsigma}{\mathbf{\Sigma}}

\newcommand{\Davg}{D_\text{avg}}
\newcommand{\mAP}{\operatorname{mAP}}

\newcommand{\G}{\mathcal{G}}
\newcommand{\V}{\mathcal{V}}
\newcommand{\edges}{\mathcal{E}}
\newcommand{\A}{\mathbf{A}}

\newcommand{\Ahat}{\mathbf{\hat{A}}}
\newcommand{\I}{\mathbf{I}}

\newcommand{\nsamples}{n_\text{samples}}
\newcommand{\npoints}{n_\text{points}}
\newcommand{\nepochs}{n_\text{epochs}}
\newcommand{\Ldistortion}{\mathcal{L}_\text{distortion}}

\newcommand{\Manify}{\texttt{Manify}}
\newcommand{\Geoopt}{\texttt{Geoopt}}
\newcommand{\sklearn}{\texttt{Scikit-Learn}}

\usepackage{tikz}
\usetikzlibrary{arrows.meta, positioning,shapes.geometric,fit,backgrounds}

\usepackage{algorithm}
\usepackage{algorithmic}
\usepackage{xparse}
\usepackage{listings}
\usepackage{enumitem}

\usepackage{lastpage}
\jmlrheading{23}{2025}{1-\pageref{LastPage}}{1/21; Revised 7/25}{9/22}{21-0000}{Chlenski, Du, Satow, Khan, and Pe'er}

\ShortHeadings{\Manify: A Python Library for Learning Non-Euclidean Representations}{Chlenski, Du, Satow, Khan, and Pe'er}
\firstpageno{1}

\begin{document}

\title{\Manify: A Python Library for Learning\\
Non-Euclidean Representations}

\author{%
    \name Philippe Chlenski
    \email pac@cs.columbia.edu \AND
    \name Kaizhu Du \email kd2814@columbia.edu \AND
    \name Dylan Satow \email dms2315@columbia.edu \AND
    \name Raiyan R. Khan \email raiyan@cs.columbia.edu \AND
    \name Itsik Pe'er \email itsik@cs.columbia.edu\\
    \addr Department of Computer Science\\ Columbia University\\ New York, NY 11227, USA
}

\editor{My editor}

\maketitle

\begin{abstract}
We present \Manify, an open-source Python library for non-Euclidean representation learning. 
Leveraging manifold learning techniques, \Manify\ provides tools for learning embeddings in (products of) non-Euclidean spaces, performing classification and regression with data that lives in such spaces, estimating the curvature of a manifold, and more. 
\Manify\ aims to advance research and applications in machine learning by offering a comprehensive suite of tools for manifold-based data analysis. 
Our source code, examples, and documentation are available at \url{https://github.com/pchlenski/manify}. 
\end{abstract}


\section{Introduction}
In recent years, there has been an increased interest in non-Euclidean representation learning. 
Embedding data in hyperbolic, spherical, or mixed-curvature product spaces can produce much more faithful embeddings (as measured by metric distortion) than traditional Euclidean methods \citep{gu_learning_2018}.
This has spurred an interest in adapting conventional machine learning methods to non-Euclidean spaces. 

To support the maximum possible number of use cases, we design the \Manify\ library around Product Manifolds (PMs): Cartesian products of hyperbolic, hyperspherical, and Euclidean ``component manifolds.''
They can model heterogeneous curvature while remaining mathematically simple; moreover, they are a superclass for all other constant-curvature spaces: $\text{Euclidean Manifolds} \subset \text{Constant-Curvature Manifolds} \subset \text{PMs}$.
The unified PM perspective naturally extends to a wide variety of learning tasks and recovers all hyperbolic/Euclidean variants of machine learning models as special cases. 
Graph Convolutional Networks (GCNs), which traditionally operate in Euclidean spaces, can be considered under this framework with a single component and used to derive other models: $\text{Logistic Regression} \subset \text{MLP} \subset \text{GCN}$.

In this paper, we present \Manify, a new open-source Python package that provides three core functions: (a) embedding graphs/distance matrices into PMs, (b) training predictors for PM-valued features, and (c) measuring curvature and signature estimation for PMs.

\section{Related Work}

There are a limited number of Python packages for working with PMs. 
The most prominent one is \Geoopt\ \citep{kochurov_geoopt_2020}, which implements Riemannian Optimization for a variety of non-Euclidean manifolds, including PMs.
\Manify\ is built on top of \Geoopt\ base classes and uses its optimizers for training.

Several research groups have explored non-Euclidean machine learning, but existing implementations are fragmented across multiple repositories, highlighting the need for a unified, maintained library like \Manify, which builds upon valuable prior work (credited in Appendix~\ref{app:code_references}) while providing consistent, modern implementations across the full spectrum of non-Euclidean algorithms.

The closest prior work to \Manify\ is \path{hyperbolic_learning_library} \citep{hypLL}, which presents a Python library for deep learning in hyperbolic space, including implementations for hyperbolic neural net layers and the Poincar\'e embedding method described in \citet{nickel_poincare_2017}. 
In contrast, our library extends to PMs, includes multiple embedding methods, and extends to non-deep learning classifiers like Decision Trees \citep{chlenski_productdt_2024}, Perceptrons, and Support Vector Machines \citep{tabaghi_linear_2021}.

\section{Library Structure}
\begin{figure}[!t]
    \centering
    \scalebox{.65}{
        \begin{tikzpicture}[
    box/.style={draw, minimum width=4.5cm, minimum height=1cm, thick},
    >=Stealth,
    node distance=0.25cm and 1.5cm,
]

\node[box] (manify) at (0,0) {\texttt{manify}};
\node[box, below=of manify] (manifolds) {\texttt{manifolds}};

\node[box, right=of manify] (curvature) {\texttt{curvature\_estimation}};
\node[box, above=of curvature] (predictors) {\texttt{predictors}};
\node[box, above=of predictors] (embedders) {\texttt{embedders}};
\node[box, below=of curvature] (clustering) {\texttt{clustering}};
\node[box, below=of clustering] (optimizers) {\texttt{optimizers}};
\node[box, below=of optimizers] (utils) {\texttt{utils}};

\draw (manify) -- (manifolds);
\draw[-] (manify) -- ++(3,0) |- (curvature);
\draw[-] (manify) -- ++(3,0) |- (predictors);
\draw[-] (manify) -- ++(3,0) |- (embedders);
\draw[-] (manify) -- ++(3,0) |- (clustering);
\draw[-] (manify) -- ++(3,0) |- (optimizers);
\draw[-] (manify) -- ++(3,0) |- (utils);

\node[right=0.2cm of curvature, text width=12cm, align=left] 
    {$\delta$-hyperbolicity, sectional curvature, etc.};
\node[right=0.2cm of predictors, text width=12cm, align=left] 
    {Product space GCNs, decision trees, perceptrons, SVMs};
\node[right=0.2cm of embedders, text width=12cm, align=left] 
    {Embedding via coordinate learning, VAEs, or Siamese nets};
\node[right=0.2cm of clustering, text width=12cm, align=left] 
    {Clustering algorithms on product manifolds};
\node[right=0.2cm of optimizers, text width=12cm, align=left] 
    {Riemannian optimization algorithms};
\node[right=0.2cm of utils, text width=12cm, align=left] 
    {Loading data, benchmarking, visuals, and other conveniences};

\end{tikzpicture}
    }
    \caption{The structure of the \Manify\ library, annotated with short descriptions of what can be found in each submodule.}
    \label{fig:library_structure}
\end{figure}
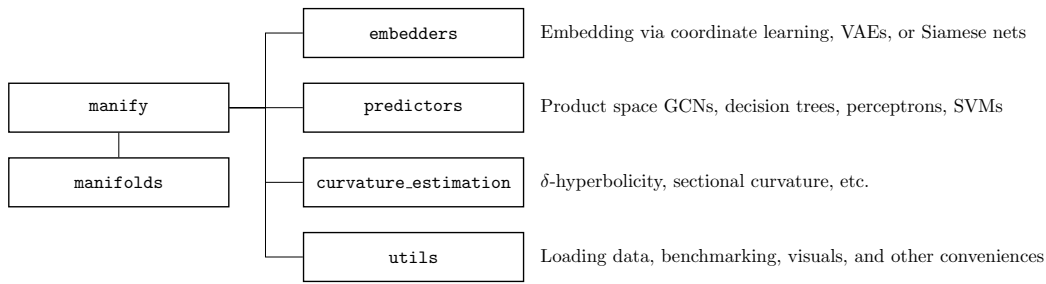
In this section, we will discuss the main features in \texttt{Manify}. 
Interested readers can find thorough mathematical details for all of the methods implemented in \Manify\ in Appendices~\ref{app:manifolds},~\ref{app:embedders},~\ref{app:predictors}, and~\ref{app:curvature}. 

\path{Manify.manifolds} consists of two classes: \path{Manifold} and \path{ProductManifold}, which are extension of their respective \Geoopt\ classes. 
The Manifold class handles single-component constant curvature Euclidean, hyperbolic, and hyperspherical manifolds. 
The ProductManifold class supports products of multiple manifolds, combining their geometric properties to create mixed-curvature representations. 
Both classes implement methods supporting key manifold operations such as distance computations and logarithmic and exponential maps.
More details are included in Appendix~\ref{app:manifolds}.
 
\path{Manify.embedders} implements methods to embed points into product manifolds on the basis of pairwise distances and/or features.
All embedders follow \sklearn\ transformer API conventions, exposing \path{fit} and \path{transform} methods.
Its implementation details can be found in Appendix~\ref{app:embedders}. 
Here are the core features in \path{embedders}:
\begin{itemize}[noitemsep,topsep=0pt]
    \item \path{Manify.embedders.coordinate_learning}: the coordinate learning algorithm proposed by \citet{gu_learning_2018}
    \item \path{Manify.embedders.siamese}: Siamese Neural Networks compatible with product space geometry, modeled off of \citet{corso_neural_2021}. 
    \item \path{Manify.embedders.vae}: the Mixed-Curvature Variational Autoencoder (VAE) proposed by \citet{skopek_mixed-curvature_2020}\\
\end{itemize}

\path{Manify.predictors} implements predictors that work on data with product manifold-valued features.
All predictors follow \sklearn\ predictor API conventions, exposing \path{fit}, \path{predict_proba}, \path{predict}, and \path{score} methods. 
Further mathematical details on these algorithms can be found in Appendix~\ref{app:predictors}. 
Here are the core features of \path{predictors}:
\begin{itemize}[noitemsep,topsep=0pt]
    \item \path{Manify.predictors.decision_tree}: the Product Space Decision Tree and Random Forest predictors proposed by \citet{chlenski_productdt_2024}
    \item \path{Manify.predictors.kappa_gcn}: the $\kappa$-GCN predictor proposed by
    \citet{bachmann_constant_2020}
    \item \path{Manify.predictors.perceptron}: Product Space Perceptron proposed by \citet{tabaghi_linear_2021}
    \item \path{Manify.predictors.svm}: the Product Space Support Vector Machine (SVM) proposed by \citet{tabaghi_linear_2021}
    \item \path{Manify.predictors.nn}: Submodule containing neural network layers, including $\kappa$-GCN layers, stereographic logits, Fermi-Dirac decoder layers, and stereographic transformer blocks \citep{cho_curve_2023}.\\
\end{itemize}

\path{Manify.curvature_estimation} provides functions for estimating the curvature of a dataset based on a graph or matrix of pairwise distances, helping users estimate what manifolds are appropriate for embedding their data. 
More details are included in Appendix~\ref{app:curvature}.
Here are the core features of \path{curvature_estimation}:
\begin{itemize}[noitemsep,topsep=0pt]
    \item \path{Manify.curvature_estimation.delta_hyperbolicity}: $\delta$- hyperbolicity estimation, as proposed by \citet{Gromov1987}. We use the efficient min-max formulation proposed by \citet{Jonckheere2010}, as seen in \citet{khrulkov}.
    \item \path{Manify.curvature_estimation.sectional_curvature}: the per-node sectional curvature estimation method from \citet{gu_learning_2018}
    \item \path{Manify.curvature_estimation.greedy_method}: the Greedy heuristic used to construct near-optimal signatures in \citet{tabaghi_linear_2021}\\
\end{itemize}

\path{Manify.clustering} provides functions for clustering points on product manifolds. Currently, it only implements the Riemannian fuzzy $K$-means algorithm proposed by \citet{yuan_riemannian_2025}.
\path{Manify.optimizers} contains Riemannian optimizers that are not presently implemented in \Geoopt.
Currently, it contains an implementation of Radan, a Riemannian variant of Adan \citep{xie_adan_2024}.
Radan is also originally proposed in \citet{yuan_riemannian_2025}.
Details for both clustering and optimizers, and their relationship to the Riemannian Fuzzy $K$-Means algorithm, are given in Appendix~\ref{app:clustering}.

\path{Manify.utils} also provides a set of convenience functions supporting typical use cases:
\begin{itemize}[noitemsep,topsep=0pt]
    \item \path{Manify.utils.benchmarks}: Tools to streamline benchmarking.
    \item \path{Manify.utils.dataloaders}: Tools for loading different datasets.
    \item \path{Manify.utils.link_prediction}: Preprocessing graphs for link prediction tasks
    \item \path{Manify.utils.visualization}: Tools for visualization 
\end{itemize}

\section{Example Usage}
\texttt{Manify} can simply be downloaded from PyPi by \texttt{pip install manify}.
This code snippet demonstrates how \Manify\ can be used to load, embed, and classify the \texttt{polblogs} dataset:

\begin{lstlisting}[language=Python, 
                    basicstyle=\ttfamily\small,
                    keywordstyle=\color{blue},
                    commentstyle=\color{green!60!black},
                    stringstyle=\color{red},
                    showstringspaces=false]
import manify
from manify.utils.dataloaders import load_hf
from sklearn.model_selection import train_test_split

# Load Polblogs graph from HuggingFace
features, dists, adj, labels = load_hf("polblogs")

# Create an S^4 x H^4 product manifold
pm = manify.ProductManifold(signature=[(1.0, 4), (-1.0, 4)])

# Learn embeddings (Gu et al (2018) method)
embedder = manify.CoordinateLearning(pm=pm)
X_embedded = embedder.fit_transform(
    X=None, D=dists, burn_in_iterations=200, training_iterations=800
)

# Train and evaluate classifier (Chlenski et al (2025) method)
X_train, X_test, y_train, y_test = train_test_split(X_embedded, labels)
model = manify.ProductSpaceDT(pm=pm, max_depth=3, task="classification")
model.fit(X_train, y_train)
print(model.score(X_test, y_test))
\end{lstlisting}

\section{Conclusion and Future Work}
We present \Manify, an open-source Python package for machine learning in mixed-curvature product manifolds. 
Although it is particularly well-suited for features (or distances) conforming to spaces of heterogeneous curvature, such as hierarchical data, networks, and certain types of biological data, it nonetheless provides the tools for carrying out machine learning methods in any of the geometries described. Future work should include:
\begin{itemize}[noitemsep,topsep=0pt]
    \item Carefully validating additional models against the results reported in the original papers (VAEs, Siamese Neural Nets, Product Space Perceptrons, Product Space SVMs, and sectional curvature estimation);
    \item Adding additional models such as product space PCA \citep{tabaghi_principal_2024});
    \item Fully aligning \Manify\ with \sklearn\ conventions;
    \item Applying \Manify\ to more datasets and workflows.
\end{itemize}


\acks{This work was funded by NSF GRFP grant DGE-2036197. We thank Jinghui Yuan for contributing the Riemannian Fuzzy K-Means algorithm to the \Manify\ library. 
}


\vskip 0.2in
\bibliography{references}

\begin{thebibliography}{34}
\providecommand{\natexlab}[1]{#1}
\providecommand{\url}[1]{\texttt{#1}}
\expandafter\ifx\csname urlstyle\endcsname\relax
  \providecommand{\doi}[1]{doi: #1}\else
  \providecommand{\doi}{doi: \begingroup \urlstyle{rm}\Url}\fi

\bibitem[Bachmann et~al.(2020)Bachmann, Bécigneul, and Ganea]{bachmann_constant_2020}
Gregor Bachmann, Gary Bécigneul, and Octavian-Eugen Ganea.
\newblock Constant {Curvature} {Graph} {Convolutional} {Networks}, May 2020.
\newblock URL \url{http://arxiv.org/abs/1911.05076}.
\newblock arXiv:1911.05076 [cs].

\bibitem[Breiman(2001)]{breiman_random_2001}
Leo Breiman.
\newblock Random forests.
\newblock \emph{Machine Learning}, 45\penalty0 (1):\penalty0 5--32, October 2001.
\newblock ISSN 1573-0565.
\newblock \doi{10.1023/A:1010933404324}.
\newblock URL \url{https://doi.org/10.1023/A:1010933404324}.

\bibitem[Breiman(2017)]{breiman_classification_2017}
Leo Breiman.
\newblock \emph{Classification and {Regression} {Trees}}.
\newblock Routledge, New York, October 2017.
\newblock ISBN 978-1-315-13947-0.
\newblock \doi{10.1201/9781315139470}.

\bibitem[Chami et~al.(2019)Chami, Ying, Ré, and Leskovec]{chami_hyperbolic_2019}
Ines Chami, Rex Ying, Christopher Ré, and Jure Leskovec.
\newblock Hyperbolic {Graph} {Convolutional} {Neural} {Networks}, October 2019.
\newblock URL \url{http://arxiv.org/abs/1910.12933}.
\newblock arXiv:1910.12933 [cs, stat].

\bibitem[Chlenski et~al.(2024)Chlenski, Turok, Moretti, and Pe'er]{chlenski_fast_2024}
Philippe Chlenski, Ethan Turok, Antonio Moretti, and Itsik Pe'er.
\newblock Fast hyperboloid decision tree algorithms, March 2024.
\newblock URL \url{http://arxiv.org/abs/2310.13841}.
\newblock arXiv:2310.13841 [cs].

\bibitem[Chlenski et~al.(2025)Chlenski, Chu, Khan, Du, Moretti, and Pe'er]{chlenski_productdt_2024}
Philippe Chlenski, Quentin Chu, Raiyan Khan, Kaizhu Du, Antonio Moretti, and Itsik Pe'er.
\newblock Mixed-curvature decision trees and random forests, October 2025.
\newblock URL \url{https://arxiv.org/pdf/2410.13879}.
\newblock arXiv:2410.13879v2 [cs].

\bibitem[Cho et~al.(2023)Cho, Cho, Park, Lee, Lee, and Lee]{cho_curve_2023}
Sungjun Cho, Seunghyuk Cho, Sungwoo Park, Hankook Lee, Honglak Lee, and Moontae Lee.
\newblock Curve {Your} {Attention}: {Mixed}-{Curvature} {Transformers} for {Graph} {Representation} {Learning}, September 2023.
\newblock URL \url{http://arxiv.org/abs/2309.04082}.
\newblock arXiv:2309.04082 [cs].

\bibitem[Corso et~al.(2021)Corso, Ying, Pándy, Velickovic, Leskovec, and Liò]{corso_neural_2021}
Gabriele Corso, Rex Ying, Michal Pándy, Petar Velickovic, Jure Leskovec, and Pietro Liò.
\newblock Neural {Distance} {Embeddings} for {Biological} {Sequences}, October 2021.
\newblock URL \url{http://arxiv.org/abs/2109.09740}.
\newblock arXiv:2109.09740 [cs, q-bio].

\bibitem[De~Sa et~al.(2018)De~Sa, Gu, Ré, and Sala]{de_sa_representation_2018}
Christopher De~Sa, Albert Gu, Christopher Ré, and Frederic Sala.
\newblock Representation {Tradeoffs} for {Hyperbolic} {Embeddings}, April 2018.
\newblock URL \url{http://arxiv.org/abs/1804.03329}.
\newblock arXiv:1804.03329 [cs, stat].

\bibitem[Deng et~al.(2009)Deng, Dong, Socher, Li, Li, and Fei-Fei]{deng2009imagenet}
Jia Deng, Wei Dong, Richard Socher, Li-Jia Li, Kai Li, and Li~Fei-Fei.
\newblock Imagenet: A large-scale hierarchical image database.
\newblock In \emph{2009 IEEE Conference on Computer Vision and Pattern Recognition}, pages 248--255. Ieee, 2009.

\bibitem[Do~Carmo(1992)]{do_carmo_riemannian_1992}
Manfredo Do~Carmo.
\newblock \emph{Riemannian {Geometry}}.
\newblock Springer US, 1992.
\newblock URL \url{https://link.springer.com/book/9780817634902}.

\bibitem[Fournier et~al.(2015)Fournier, Ismail, and Vigneron]{FOURNIER2015576}
Hervé Fournier, Anas Ismail, and Antoine Vigneron.
\newblock Computing the gromov hyperbolicity of a discrete metric space.
\newblock \emph{Information Processing Letters}, 115\penalty0 (6):\penalty0 576--579, 2015.
\newblock ISSN 0020-0190.
\newblock \doi{https://doi.org/10.1016/j.ipl.2015.02.002}.
\newblock URL \url{https://www.sciencedirect.com/science/article/pii/S0020019015000198}.

\bibitem[Ganea et~al.(2018)Ganea, Bécigneul, and Hofmann]{ganea_hyperbolic_2018}
Octavian-Eugen Ganea, Gary Bécigneul, and Thomas Hofmann.
\newblock Hyperbolic {Neural} {Networks}, June 2018.
\newblock URL \url{http://arxiv.org/abs/1805.09112}.
\newblock arXiv:1805.09112 [cs].

\bibitem[Gromov(1987)]{Gromov1987}
Mikhail Gromov.
\newblock Hyperbolic groups.
\newblock In S.~M. Gersten, editor, \emph{Essays in Group Theory}, volume~8 of \emph{Mathematical Sciences Research Institute Publications}, pages 75--263. Springer, 1987.
\newblock \doi{10.1007/978-1-4613-9586-7_3}.

\bibitem[Gu et~al.(2018)Gu, Sala, Gunel, and Ré]{gu_learning_2018}
Albert Gu, Frederic Sala, Beliz Gunel, and Christopher Ré.
\newblock Learning {Mixed}-{Curvature} {Representations} in {Product} {Spaces}.
\newblock September 2018.
\newblock URL \url{https://openreview.net/forum?id=HJxeWnCcF7}.

\bibitem[He et~al.(2016)He, Zhang, Ren, and Sun]{he2016deep}
Kaiming He, Xiangyu Zhang, Shaoqing Ren, and Jian Sun.
\newblock Deep residual learning for image recognition.
\newblock In \emph{Proceedings of the IEEE conference on computer vision and pattern recognition}, pages 770--778, 2016.

\bibitem[Jonckheere et~al.(2010)Jonckheere, Lohsoonthorn, and Bonahon]{Jonckheere2010}
Edmond~A. Jonckheere, Patrice Lohsoonthorn, and Fabrice Bonahon.
\newblock Scaled gromov four-point condition for network graph curvature computation.
\newblock \emph{Internet Mathematics}, 7\penalty0 (3):\penalty0 137--177, 2010.
\newblock \doi{10.1080/15427951.2010.10129177}.

\bibitem[Khrulkov et~al.(2019)Khrulkov, Mirvakhabova, Ustinova, Oseledets, and Lempitsky]{khrulkov}
Valentin Khrulkov, Leyla Mirvakhabova, Evgeniya Ustinova, Ivan Oseledets, and Victor Lempitsky.
\newblock Hyperbolic image embeddings, April 2019.
\newblock URL \url{https://arxiv.org/abs/1904.02239}.
\newblock arXiv:1904.02239 [cs.CV].

\bibitem[Kingma and Welling(2013)]{kingma2014vae}
Diederik Kingma and Max Welling.
\newblock Auto-encoding variational bayes, December 2013.
\newblock URL \url{https://arxiv.org/abs/1312.6114}.
\newblock arXiv:1312.6114 [stats.ML].

\bibitem[Kochurov et~al.(2020)Kochurov, Karimov, and Kozlukov]{kochurov_geoopt_2020}
Max Kochurov, Rasul Karimov, and Serge Kozlukov.
\newblock Geoopt: {Riemannian} {Optimization} in {PyTorch}, July 2020.
\newblock URL \url{http://arxiv.org/abs/2005.02819}.
\newblock arXiv:2005.02819 [cs].

\bibitem[Krioukov et~al.(2010)Krioukov, Papadopoulos, Kitsak, Vahdat, and Boguñá]{krioukov_hyperbolic_2010}
Dmitri Krioukov, Fragkiskos Papadopoulos, Maksim Kitsak, Amin Vahdat, and Marián Boguñá.
\newblock Hyperbolic geometry of complex networks.
\newblock \emph{Physical Review E}, 82\penalty0 (3):\penalty0 036106, September 2010.
\newblock \doi{10.1103/PhysRevE.82.036106}.
\newblock URL \url{https://link.aps.org/doi/10.1103/PhysRevE.82.036106}.
\newblock Publisher: American Physical Society.

\bibitem[Liu et~al.(2019)Liu, Nickel, and Kiela]{liu_hyperbolic_2019}
Qi~Liu, Maximilian Nickel, and Douwe Kiela.
\newblock Hyperbolic {Graph} {Neural} {Networks}, October 2019.
\newblock URL \url{http://arxiv.org/abs/1910.12892}.
\newblock arXiv:1910.12892 [cs, stat].

\bibitem[Nagano et~al.(2019)Nagano, Yamaguchi, Fujita, and Koyama]{nagano_wrapped_2019}
Yoshihiro Nagano, Shoichiro Yamaguchi, Yasuhiro Fujita, and Masanori Koyama.
\newblock A {Wrapped} {Normal} {Distribution} on {Hyperbolic} {Space} for {Gradient}-{Based} {Learning}, May 2019.
\newblock URL \url{http://arxiv.org/abs/1902.02992}.
\newblock arXiv:1902.02992 [cs, stat].

\bibitem[Nickel and Kiela(2017)]{nickel_poincare_2017}
Maximilian Nickel and Douwe Kiela.
\newblock Poincaré {Embeddings} for {Learning} {Hierarchical} {Representations}, May 2017.
\newblock URL \url{http://arxiv.org/abs/1705.08039}.
\newblock arXiv:1705.08039.

\bibitem[Pedregosa et~al.(2011)Pedregosa, Varoquaux, Gramfort, Michel, Thirion, Grisel, Blondel, Prettenhofer, Weiss, Dubourg, Vanderplas, Passos, Cournapeau, Brucher, Perrot, and Duchesnay]{pedregosa_scikit-learn_2011}
Fabian Pedregosa, Gaël Varoquaux, Alexandre Gramfort, Vincent Michel, Bertrand Thirion, Olivier Grisel, Mathieu Blondel, Peter Prettenhofer, Ron Weiss, Vincent Dubourg, Jake Vanderplas, Alexandre Passos, David Cournapeau, Matthieu Brucher, Matthieu Perrot, and Edouard Duchesnay.
\newblock Scikit-learn: {Machine} {Learning} in {Python}.
\newblock \emph{Journal of Machine Learning Research}, 12\penalty0 (85):\penalty0 2825--2830, 2011.
\newblock ISSN 1533-7928.
\newblock URL \url{http://jmlr.org/papers/v12/pedregosa11a.html}.

\bibitem[Simonyan and Zisserman(2015)]{simonyan2015very}
Karen Simonyan and Andrew Zisserman.
\newblock Very deep convolutional networks for large-scale image recognition.
\newblock \emph{arXiv preprint arXiv:1409.1556}, 2015.

\bibitem[Skopek et~al.(2020)Skopek, Ganea, and Bécigneul]{skopek_mixed-curvature_2020}
Ondrej Skopek, Octavian-Eugen Ganea, and Gary Bécigneul.
\newblock Mixed-curvature {Variational} {Autoencoders}, February 2020.
\newblock URL \url{http://arxiv.org/abs/1911.08411}.
\newblock arXiv:1911.08411 [cs, stat].

\bibitem[Spengler et~al.(2023)Spengler, Wirth, and Mettes]{hypLL}
Max Spengler, Philipp Wirth, and Pascal. Mettes.
\newblock Hypll: The hyperbolic learning library, June 2023.
\newblock URL \url{https://arXiv:2306.06154}.
\newblock arXiv.2306.06154 [cs.LG].

\bibitem[Szegedy et~al.(2016)Szegedy, Vanhoucke, Ioffe, Shlens, and Wojna]{szegedy2016rethinking}
Christian Szegedy, Vincent Vanhoucke, Sergey Ioffe, Jon Shlens, and Zbigniew Wojna.
\newblock Rethinking the inception architecture for computer vision.
\newblock In \emph{Proceedings of the IEEE conference on computer vision and pattern recognition}, pages 2818--2826, 2016.

\bibitem[Tabaghi et~al.(2021)Tabaghi, Pan, Chien, Peng, and Milenkovic]{tabaghi_linear_2021}
Puoya Tabaghi, Chao Pan, Eli Chien, Jianhao Peng, and Olgica Milenkovic.
\newblock Linear {Classifiers} in {Product} {Space} {Forms}, February 2021.
\newblock URL \url{http://arxiv.org/abs/2102.10204}.
\newblock arXiv:2102.10204 [cs, stat] version: 1.

\bibitem[Tabaghi et~al.(2024)Tabaghi, Khanzadeh, Wang, and Mirarab]{tabaghi_principal_2024}
Puoya Tabaghi, Michael Khanzadeh, Yusu Wang, and Sivash Mirarab.
\newblock Principal {Component} {Analysis} in {Space} {Forms}, July 2024.
\newblock URL \url{http://arxiv.org/abs/2301.02750}.
\newblock arXiv:2301.02750 [cs, eess, math, stat].

\bibitem[Wu et~al.(2019)Wu, Zhang, Jr., Fifty, Yu, and Weinberger]{wu_simplifying_2019}
Felix Wu, Tianyi Zhang, Amauri H.~Souza Jr., Christopher Fifty, Tao Yu, and Kilian~Q. Weinberger.
\newblock Simplifying graph convolutional networks.
\newblock \emph{CoRR}, abs/1902.07153, 2019.
\newblock URL \url{http://arxiv.org/abs/1902.07153}.

\bibitem[Xie et~al.(2024)Xie, Zhou, Li, Lin, and Yan]{xie_adan_2024}
Xingyu Xie, Pan Zhou, Huan Li, Zhouchen Lin, and Shuicheng Yan.
\newblock Adan: {Adaptive} {Nesterov} {Momentum} {Algorithm} for {Faster} {Optimizing} {Deep} {Models}, November 2024.
\newblock URL \url{http://arxiv.org/abs/2208.06677}.
\newblock arXiv:2208.06677 [cs].

\bibitem[Yuan et~al.(2025)Yuan, Lin, Nie, and Li]{yuan_riemannian_2025}
Jinghui Yuan, Zhuo Lin, Feiping Nie, and Xuelong Li.
\newblock Riemannian {Fuzzy} {K}-{Means}.
\newblock \emph{arXiv}, 2025.
\newblock URL \url{https://openreview.net/forum?id=9VmOgMN4Ie}.

\end{thebibliography}

\newpage

\appendix
\section{Code Referenced in this Work}
\label{app:code_references}
While working on \Manify, we took inspiration from a variety of existing codebases containing implementations of various components of things we have put into \Manify. In Table~\ref{tab:repo_credits}, we credit the authors of these repos for their work:
\begin{table}[h]
    \centering
    \caption{This table enumerates and cites any online code resources we consulted during the creation of \Manify.}
    \begin{small}
    \begin{tabular}{p{6cm}p{5cm}p{2.5cm}}
        \toprule
        \textbf{Repository URL} & \textbf{Purpose} & \textbf{Citation}\\
        \midrule
        \url{https://openreview.net/attachment?id=AN5uo4ByWH&name=supplementary_material} & Code for $\kappa$-Stereographic Logits in Product Spaces & \citet{cho_curve_2023}\\
        \midrule
        \url{https://github.com/HazyResearch/hyperbolics/tree/master/products} & Variant of coordinate learning; sectional curvature & \citet{gu_learning_2018, de_sa_representation_2018}\\
        \midrule
        \url{https://github.com/thupchnsky/product-space-linear-classifiers} & Product Space Perceptron and SVM & \citet{tabaghi_linear_2021}\\
        \midrule
        \url{https://github.com/oskopek/mvae} & Mixed-Curvature VAEs & \citet{skopek_mixed-curvature_2020}
        \\
        \midrule
        \url{https://github.com/leymir/hyperbolic-image-embeddings} & Delta-hyperbolicity & \citet{khrulkov, FOURNIER2015576}\\
        \midrule
        \url{https://github.com/sail-sg/Adan} & Adan & \citet{xie_adan_2024}\\
        \midrule
        \url{https://github.com/Yuan-Jinghui/Riemannian-Fuzzy-K-Means} & Riemannian Fuzzy $K$-Means & \citet{yuan_riemannian_2025}\\
        \bottomrule
    \end{tabular}
    \end{small}
    \label{tab:repo_credits}
\end{table}

\vfill
\pagebreak
\section{List of Symbols}
\label{app:list_of_symbols}

\begin{table}[h]
    \centering
    \caption{Glossary of variables and symbols used in this paper.}
    \label{tab:symbols}
    \begin{small}
    \begin{tabular}{r c p{10cm}}
        \toprule
        \textbf{Domain} & \textbf{Symbol} & \textbf{Used for} \\
        \midrule
        Constants 
            & $N$ & Number of samples \\
            & $d$ & Number of dimensions \\
            & $\kappa$ & Curvature of a manifold \\
        \midrule
        
        Matrices 
            & $\X$ & Matrix of samples in $\R{n \times d}$ \\
            & $\W$ & Weight matrix in $\R{d \times e}$\\
            & $\y$ & Vector of labels in $\R{n}$ \\
        \midrule
        
        Manifolds 
            & $\M$ & Riemannian manifold \\
            & $\P$ & Product manifold\\
            & $\p$ & Point in a manifold, $\p \in \M$ \\
            & $\T[\p]{\M}$ & The tangent space of point $\p$ in $\M$ (a vector space) \\
            & $\v$ & A vector in a tangent plane \\
            & $g$ & Riemannian metric defining an inner product on $\T[\p]{\M}$ \\
            & $\langle \u, \v \rangle_\M$ & Inner product of $\u, \v \in \T[\p]{\M}$\\
            & $\|\v\|_\M$ & Manifold-appropriate norm of $\v$ in $\T[\p]{\M}$\\
            & $\delta_\M(\x, \y)$ & Geodesic distance between $\x, \y \in \M$ \\
            & $\E{d}$ & $d$-dimensional Euclidean space with curvature $\kappa = 0$ \\
            & $\S{d}_\kappa$ & $d$-dimensional spherical space with curvature $\kappa > 0$ \\
            & $\H{d}_\kappa$ & $d$-dimensional hyperbolic space with curvature $\kappa < 0$ \\
            & $\PT{\x}{\y}{\v}$ & Parallel transport of $\v \in \T[\x]{\M}$ to $\T[\y]{\M}$ \\
            & $\logmap{\x}{\p}$ & Logarithmic map of $\p \in \M$ to $\T[\x]{\M}$ \\
            & $\expmap{\x}{\v}$ & Exponential map of $\v \in \T[\x]{\M}$ to $\M$ \\
            & $\origin$ & The origin of $\M$\\
        \midrule
        
        $\kappa$-stereo-
            & $\st{d}{\kappa}$ & $d$-dimensional $\kappa$-stereographic model \\
        graphic model 
            & $\u \oplus_{\kappa} \v$ & $\kappa$-stereographic addition \\
            & $c \otimes_{\kappa} \v$ & $\kappa$-stereographic scaling \\
            & $\x \otimes_\kappa \W$ & $\kappa$-right vector-matrix-multiplication of $\W \in \R{d \times e}, \x \in \st{d}{\kappa}$ \\
            & $\X \otimes_\kappa \W$ & $\kappa$-right-matrix-multiplication of $\W \in \R{d \times e}, \X \in \st{n \times d}{\kappa}$ \\
            & $\mathbf{a}\boxtimes_\kappa \X$ & $\kappa$-weighted midpoint, $\mathbf{a} \in \R{n}, \X \in \st{n \times d}{\kappa}$ \\
            & $\A\boxtimes_\kappa \X$ & $\kappa$-left-matrix-multiplication of $\A \in \R{n \times n}, \X \in \st{n \times d}{\kappa}$ \\
        \midrule

        Probability
            & $\boldmu$ & Mean of a distribution, $\boldmu \in \R{d}$\\
            & $\boldsigma$ & Covariance matrix of a distribution, $\boldsigma \in \R{d \times d}$\\
            & $\N(\boldmu, \boldsigma)$ & Normal distribution\\
            & $\WN(\boldmu, \boldsigma)$ & Wrapped normal distribution\\
        \midrule
        
        Graphs 
            & $\G$ & A graph: nodes and edges $(\V, \edges)$ \\
            & $\A$ & Adjacency matrix for a graph \\
        \midrule
        
        Metrics 
            & $\Davg$ & Average distortion in an embedding \\
            & $\mAP$ & Mean average precision of a graph embedding\\
        \bottomrule
    \end{tabular}
    \end{small}
\end{table}

\vfill
\pagebreak
\section{Mathematical Details: Manifolds }
\label{app:manifolds}
\subsection{Core Riemannian Manifolds}
In this section, we will review fundamental concepts in Riemannian manifolds \citep{do_carmo_riemannian_1992}.
Let $\M$ be a smooth manifold, $\p \in \M$ a point, and $\T[\p]{\M}$ the tangent space at $\p$.
$\M$ is considered a Riemannian manifold if  $\M$ is equipped with a Riemannian metric $g$.
The Riemannian metric represents the choice of inner product for each tangent space of the manifold, defining key geometric notions such as angle and length. 
The shortest-distance paths are termed geodesics. 

Curvature measures and describes how much a geometric object deviates from a flat plane. 
If $\M$ is a constant-curvature manifold, it can be described by its curvature $\kappa$ and dimensionality $d$. 
The sign of $\kappa$ indicates if a single component manifold is hyperbolic, Euclidean, or hyperspherical.

\subsubsection{Hyperbolic Space}
Hyperbolic space is defined by its constantly negative curvature. 
Hyperbolic geometry is represented in \Manify\ by the hyperboloid model, which is also known as the Lorentz model. 
A $d$-dimensional hyperbolic space is embedded inside an ambient $(d + 1)$-dimensional Minkowski space. 
Minkowski space is a metric space equipped with the Minkowski inner product (for two vectors $\u, \v \in \R{d+1}$):
\begin{equation}
    {\langle \u, \v \rangle}_\H{} = -u_0v_0 + \sum_{i=1}^{d+1} u_iv_i\label{eq:minkowski_inner_product}
\end{equation}
We thus denote the hyperbolic manifold as a set of points with constant Minkowski norm:
\begin{equation}
    \mathbb{H}^{d}_\kappa = \left\{
        \textbf{x}\in \mathbb{R}^{d+1}: {\langle \textbf{x},\textbf{x} \rangle}_\H{} = -\frac{1}{\kappa},\ x_0>0 
    \right\},
    \label{minkowski-norm}
\end{equation}
 where the Euclidean tangent space at point $\p$ in the hyperboloid model and the distance metric in the hyperboloid model is 
 \begin{equation}
    \delta(\x, \y)_\H{} = \frac{
        \cosh^{-1} (-\kappa {\langle \x, \y \rangle}_\H{})
    }{
        \sqrt{-\kappa}
    }.
    \label{hyp_dist}
 \end{equation}

\subsubsection{Euclidean Space}
Euclidean space is defined by its constant zero curvature. 
Although vectors spaces behave as Euclidean space, we use the symbol $\E{d}$ to denote the $d$-dimensional Euclidean space, so that we can differentiate it from other uses of $\R{d}$ (e.g. in defining Minkowski space above).
The inner product of two vectors $\u$ and $\v$ is defined as its dot product:
\begin{equation}
    \langle \mathbf{u}, \mathbf{v} \rangle_\E{} = u_0 v_0 + u_1 v_1 + \ldots + u_2 v_2. 
    \label{dot_product}
\end{equation}
The norm of a vector \(\mathbf{u}\) is given by the $\ell_2$ norm:
\begin{equation}
    \|\mathbf{u}\|_\E{} = \sqrt{\langle \mathbf{u}, \mathbf{u} \rangle}_\E{}. 
    \label{l2_norm}
\end{equation}
The Euclidean distance between two vectors $\x, \y \in \E{}$ is defined as:
\begin{equation}
    \delta_\E{}(\x, \y) = \|\mathbf{u} - \mathbf{v}\|_\E{}. 
    \label{euc_dist}
\end{equation}

\subsubsection{Hyperspherical Space}
Hyperspherical space is defined by its constant positive curvature. 
It has the same inner products as Euclidean space. 
This space $\S{d}_\kappa$ is most easily modeled and defined in:
\begin{equation}
    \S{d}_\kappa = \left\{ 
        \x \in \R{d+1} \suchthat \|\x\|_\E{} = \frac{1}{\sqrt{\kappa}}
    \right\},
    \label{hyperspherical_space}
\end{equation}
with the following distance function for $\x, \y \in \S{}$:
\begin{equation}
    \delta_\S{}(\u, \v) = \frac{\cos^{-1}(\kappa^2\dot{\u}{\v}_\E{})}{\kappa}.
    \label{dist_spherical}
\end{equation}

\subsection{The $\kappa$-Stereographic Model}
\subsubsection{Manifold Definitions}
For the $\kappa$-GCN, we further need to describe the \(\kappa\)-stereographic model of constant-curvature spaces. 
For a curvature $\kappa \in \R{}$ and a dimension $d \geq 2$, the $\kappa$-sterographic model in $d$ dimensions $\st{d}{\kappa}$ is defined as
\begin{equation}
    \st{d}{\kappa} = \{\x \in \R{d} \suchthat -\kappa \langle\x, \x\rangle_\E{} < 1\}
\label{k_stero_model}
\end{equation}
with the Riemannian metric $g_\x^\kappa$ defined in terms of a conformal factor at $\x$, $ \lambda_x^\kappa$
\begin{align}
    g_\x^\kappa &= (\lambda_\x^\kappa)^2 \I, \text{ where}\\
    \lambda_\x^\kappa &= \frac{2}{1 + \kappa \langle \x, \x \rangle_\E{}}.
\label{metric}
\end{align}

\subsubsection{Stereographic Projections}
For a point $\mathbf{x} \in \mathcal{M}_\kappa$, where $\mathcal{M}_\kappa$ is one of $\mathbb{H}^d_\kappa, \mathbb{E}^d,$ or $\mathbb{S}^d_\kappa$ depending on $\kappa$, we can rewrite $\mathbf{x} = (x_0, \mathbf{x}_\text{rest}) \in \mathbb{R}^{d+1}$ where $x_0 \in \mathbb{R}$ is the first coordinate and $\mathbf{x}_\text{rest} \in \mathbb{R}^d$ contains the remaining coordinates.

Then the stereographic projection $\rho_\kappa: \mathcal{M}_\kappa \to \st{d}{\kappa}$ and its inverse $\rho_\kappa^{-1}: \st{d}{\kappa} \to \mathcal{M}_\kappa$ are defined as:
\begin{align}
    \rho_\kappa((x_0, \mathbf{x}_\text{rest})) &= \frac{\mathbf{x}_\text{rest}}{1 + \sqrt{|\kappa|}x_0},\\
    \rho_\kappa^{-1}(\mathbf{x}') &= \left(
        \frac{1 - \kappa\|\mathbf{x}'\|^2}{\sqrt{|\kappa|}(1 + \kappa\|\mathbf{x}'\|^2)}, \frac{2\mathbf{x}'}{1 + \kappa\|\mathbf{x}'\|^2}
    \right).
\end{align}
These projections establish an isomorphism between the $\kappa$-stereographic model and the sphere, hyperbolic space, or Euclidean space depending on the sign of $\kappa$:
\begin{itemize}[noitemsep,topsep=0pt]
    \item For $\kappa > 0$: Maps between $\st{d}{\kappa}$ and the sphere $\mathbb{S}^{d}_{\kappa}$ with positive curvature
    \item For $\kappa = 0$: Simplifies to the identity mapping on Euclidean space $\mathbb{E}^{d}$
    \item For $\kappa < 0$: Maps between $\st{d}{\kappa}$ and hyperbolic space $\mathbb{H}^{d}_{\kappa}$ with negative curvature
\end{itemize}

The stereographic projection preserves angles (conformal) but not distances. This allows us to work with the $\kappa$-stereographic model as a unified representation of spaces with different curvatures.

\subsubsection{Gyrovector Operations}
An advantage of the $\kappa$-stereographic model is that $\st{d}{\kappa}$ makes up a gyrovector space, which enables us to compute closed forms for operations such as addition, scalar multiplication, and even matrix multiplication directly in $\st{d}{\kappa}$.

\paragraph{Addition.} 
We define the \(\kappa\)-addition in the \(\kappa\)-stereographic model for \(\mathbf{x}, \mathbf{y} \in \mathfrak{st}^d_{\kappa}\) by:
\begin{equation}
    \mathbf{x} \oplus_{\kappa} \mathbf{y} = \frac{(1 - 2\kappa \mathbf{x}^\top \mathbf{y} - \kappa \|\mathbf{y}\|^2) \mathbf{x} + (1 + \kappa \|\mathbf{x}\|^2) \mathbf{y}}{1 - 2\kappa \mathbf{x}^\top \mathbf{y} + \kappa^2 \|\mathbf{x}\|^2 \|\mathbf{y}\|^2}.
\label{kappa_add}
\end{equation}

\paragraph{Scalar Multiplication.} 
For \(s \in \mathbb{R}\) and \(\mathbf{x} \in \mathfrak{st}^d_{\kappa}\),  \(\kappa\)-scaling is defined as:
\begin{equation}
    s \otimes_{\kappa} \mathbf{x} = \tan_{\kappa} (s \cdot \tan^{-1}_{\kappa} \|\mathbf{x}\|) \frac{\mathbf{x}}{\|\mathbf{x}\|} \in \mathfrak{st}^d_{\kappa},
\label{kappa_scale}
\end{equation}
where $\tan_\kappa$ = $\tanh$ if $\kappa < 0$, and $\tan$ otherwise.

\paragraph{Right-Multiplication.}
Before we can define $\kappa$-right matrix multiplication, we must define the $\kappa$-right vector-matrix multiplication $\x \otimes_\kappa \W$ for $\x \in \st{d}{\kappa},\ \W \in \R{d \times d}$:
\begin{align}
    \x \otimes_\kappa \W
    &= \exp_{\mathbf{0}}(\log_{\mathbf{0}}(\x)\W)\\
    &= \tan_\kappa \left( 
        \frac{\|(\x\W)\|}{\|\x\|} \tan_\kappa^{-1} (\|\x\|) 
    \right) \frac{(\x\W)}{\|(\x\W)\|}.
\label{right_mul}
\end{align}

The $\kappa$-right matrix-matrix multiplication $\X \otimes_\kappa \W$ for $\X \in \st{n \times d}{\kappa},\ \W \in \R{d \times d}$ is just the $\kappa$-right vector-matrix multiplication applied row-wise.
For clarity, we can explicitly represent the matrix $\X$ as a stack of row vectors where each row vector $\mathbf{x_i} \in \st{d}{\kappa}$, and express the $\kappa$-right matrix multiplication as:

\begin{equation}
    \X = 
    \begin{pmatrix} 
        \text{---} & \mathbf{x_1} & \text{---} \\
        \text{---} & \mathbf{x_2} & \text{---}\\
         & \vdots \\
        \text{---} & \mathbf{x_n} & \text{---}
    \end{pmatrix}, \quad
    \X \otimes_\kappa \W =
    \begin{pmatrix} 
        \text{---} & \mathbf{\x_1} \otimes_\kappa \W & \text{---} \\
        \text{---} & \mathbf{\x_2} \otimes_\kappa \W & \text{---}\\
        & \vdots \\
        \text{---} & \mathbf{\x_n} \otimes_\kappa \W & \text{---}
    \end{pmatrix}
    \label{eq:rewrite_X}\\
\end{equation}

\paragraph{Left-Multiplication.}
Before we can describe $\kappa$-left matrix multiplication, we need to establish the definition of a weighted gyromidpoint.
For a set of points $\X \in \st{n \times d}{\kappa}$ and a vector of weights $\mathbf{a} \in \R{n}$,
the weighted gyromidpoint in $\st{d}{\kappa}$ has a closed-form expression:
\begin{align}
    \mathbf{a} \boxtimes_\kappa \X &=
    \frac{1}{2} \otimes_\kappa
        \sum_{i=1}^n \frac{
            a_i\lambda^\kappa_\mathbf{x_i}
        }{
            \sum_{j=1}^n a_j (\lambda^\kappa_\mathbf{x_j} - 1)
        } \mathbf{x_i}.
\end{align}
Now, for matrices $\A \in \R{n \times n},\ \X \in \R{n \times d}$, we can now define the row-wise \(\kappa\)-left-matrix-multiplication.
Rewriting $\A$ as a stack of column vectors, we have:
\begin{equation}
    \A = \begin{pmatrix}
        \vert & \vert & &\vert\\
        \mathbf{a_1} & \mathbf{a_2} & \ldots & \mathbf{a_3}\\
        \vert & \vert & & \vert
    \end{pmatrix}, \quad
    \A \boxtimes_\kappa \X = \begin{pmatrix}
        \text{---} &\mathbf{a_1} \boxtimes_\kappa \X & \text{---} \\
        \text{---} & \mathbf{a_2} \boxtimes_\kappa \X & \text{---}\\
        & \vdots\\
        \text{---} &\mathbf{a_n} \boxtimes_\kappa \X & \text{---}
    \end{pmatrix},
\end{equation}.

\subsubsection{Manifold Operations in $\st{d}{\kappa}$}
Although \Manify\ does not make us of the following operations in $\st{d}{\kappa}$, they are inherited from \Geoopt's \texttt{Stereographic} manifold class and therefore usable.
We include them for completeness:
\begin{align}
    \delta_{\st{d}{\kappa}}(\x, \y) &= 2|\kappa|^{-1/2} \tan_{\kappa}^{-1} | -\x \oplus_{\kappa} \y |\\
    \gamma_{\x\to\y}(t) &= \x \oplus_{\kappa} (t \otimes_{\kappa} (-\x \oplus_{\kappa} \y))\\
    \expmap{x}{\st{d}{\kappa}}(\v) &= \x \oplus_{\kappa} \left( \tan_{\kappa}\left(|\kappa|^{\frac{1}{2}}\frac{\lambda_{\x}^{\kappa}|\v|}{2}\right) \frac{\v}{|\v|} \right)\\
    \logmap{x}{\st{d}{\kappa}}(\y) &= \frac{2|\kappa|^{-\frac{1}{2}}}{\lambda_{\x}^{\kappa}} \tan_{\kappa}^{-1} | -\x \oplus_{\kappa} \y| \frac{-\x \oplus_{\kappa} \y}{| -\x \oplus_{\kappa} \y |}
\end{align}

\subsection{Product Manifolds}
We follow the definition of product manifolds from \citet{gu_learning_2018}.
Formally, we define product manifold as the Cartesian product of one or more Euclidean, hyperbolic, and spherical manifolds
\begin{equation}
    \P = \prod_{i=1}^{n} \S{s_i}_{K_i} \times \prod_{j=1}^{m} \H{h_j}_{K'_j} \times \E{d}.
    \label{cartesian_product}
\end{equation}

We describe the product manifold in its signature. 
The signature includes the number of component spaces, each's dimensions, and curvature. 
Product manifold is decomposable to the individuals manifolds, or component manifolds. 
If each $\M_i$ is equipped with metric $g_i$, then distance in $\P$ decompose as the $\ell_2$ norm of the distances in each of the component manifolds:
\begin{equation}
    \delta_\P(\u, \v) = \sqrt{ \sum_{\M \in \P} \delta_{\M}(\u_i, \v_i)^2 },
    \label{dist_prod}
\end{equation} 
where $i$ represents each individual component manifold. 
Points are decomposable in that all $\p \in \P$ have coordinates $\p = (\mathbf{p_1}, \dots, \mathbf{p_k})$ where $\mathbf{p_i} \in \M_i$. 
The total number of dimensions can be decomposed into $\sum_{i}^n s_i + \sum_{j}^m h_j + d$. 
Similarly, the tangent vectors $\v \in \T[\p]{\P}$ decompose as $(\mathbf{v_1}, \dots, \mathbf{v_k})$ with $\mathbf{v_i} \in \T[\mathbf{p_i}]\M_i$. 

All operations that we rely on in other work: parallel transport, exponential and logarithmic maps, and gyrovector operations, can be done in product manifolds by applying applying these operations componentwise to the factorized embeddings, then concatenating them back together.
This ensures that product manifolds do not lose expressiveness compared to their components.

\subsection{Sampling}
\subsubsection{Wrapped Normal Distribution}
\Manify's \texttt{Manifold} and \texttt{ProductManifold} classes are equipped with a \texttt{sample} method, which relies on the wrapped normal distribution $\WN$. 
We use the formulation of the wrapped normal distribution introduced for $\H{}$ by \citet{nagano_wrapped_2019} and extended to $\S{}$ and $\st{d}{\kappa}$ by \citet{skopek_mixed-curvature_2020}.

To sample from the wrapped normal distribution $\WN$, we first sample a $\v' \in \R{d}$ from a normal distribution $\N(\mathbf{0}, \boldsigma)$; then, we convert $\v'$ to $\v \in \T[\origin]{\M}$, where the first dimension of all vectors is constant, by pretending a zero; then, we move $\v$ to $\u \in \T[\boldmu]{\M}$, where $\mu \in \M$ is the desired mean, via parallel transport; and deposit $\u$ onto $\M$ via the exponential map:
\begin{align}
    \v' &\sim \mathcal{N}(\mathbf{0}, \boldsigma)\\
    \v &= (0, \v')\\
    \mathbf{u} &= \mathrm{PT}_{\boldsymbol{\mu}_0 \to \boldsymbol{\mu}}^K (\mathbf{v})\\
    \mathbf{z} &= \expmap{\x}{\u}.
\label{eq:wrapped_normal}
\end{align}

To evaluate (log-)likelihoods in $\WN$, we exploit the fact that each of the operations defined above---parallel transport and the exponential map---has a corresponding inverse to transport points from $\M$ back to $\T[\origin]\M$, and then evaluate the log-likelihood under $\N$, adjusting for the change of coordinates via the log-determinant:
\begin{align}
    \u &= \logmap{\boldmu}{\mathbf{z}}\\
    \v &= \PT{\boldmu}{\mathbf{\mu_0}}{\u}\\
    \v' &= (v_2, v_3, \ldots, v_{d+1})\\
    \log \WN(\mathbf{z}; \boldmu, \boldsigma) &= 
    \log \N(\v; 0, \boldsigma) 
    - \log \det \left( 
        \frac{\partial \expmap{\boldmu}{\u}}{\partial \u} \cdot
        \frac{\partial \PT{\mu_0}{\mu}{\v}}{\partial \v}
    \right)\\
    &= \log \N(\v; 0, \boldsigma) - \left(
        \frac{
            \sinh(\langle \u, \u \rangle_\H{})
        }{
            \langle \u, \u \rangle_\H{}
        }
    \right)^{d-1}.
    \label{eq:wn_final_log_likelihood}
\end{align}


\subsubsection{Gaussian Mixtures for Classification and Regression}

Alongside \texttt{sample}, we include the less conventional \texttt{gaussian\_mixture} method to easily generate synthetic classification and regression datasets.
We follow the formulation of Gaussian mixtures introduced for $\H{}$ in \citet{chlenski_fast_2024} and extended to $\S{}$ and $\P$ in \citet{chlenski_productdt_2024}.
Unlike the previous implementations, our Gaussian mixtures can also be defined for $\st{d}{\kappa}$.

\paragraph{Features.}
Our approach to generating Gaussian mixtures on manifolds follows a principled process. First, we assign $n_\text{samples}$ samples to $n_\text{clusters}$ clusters by generating cluster probabilities $\mathbf{p_{norm}}$ from normalized uniform random values, then sampling categorical cluster assignments $\mathbf{c}$ from these probabilities:
\begin{align}
    \mathbf{p_{raw}} &= \langle p_0, p_1, \ldots, p_{n_\text{clusters}-1} \rangle, \quad p_i \sim \text{Uniform}(0, 1)\\
    \mathbf{p_{norm}} &= \frac{\mathbf{p_{raw}}}{ \sum_{i = 0}^{n-1} p_i}\\
    \mathbf{c} &= \langle c_0, c_1, \ldots c_{n_\text{samples}-1} \rangle, \quad c_i \sim \text{Categorical}(n, \mathbf{p_{norm}})
\end{align}

Next, we sample our class means matrix $M \in \M^{n_\text{clusters} \times d}$:
\begin{equation}
    \mathbf{M} = \begin{pmatrix}
        \text{---} & \mathbf{m_1} & \text{---}\\
        \text{---} & \mathbf{m_2} & \text{---}\\
        & \vdots\\
        \text{---} & \mathbf{m_m} & \text{---}
    \end{pmatrix},\quad \mathbf{m_i} \sim \WN\left(0, \sqrt{|\kappa|\I}\right).
\end{equation}
For each cluster $i$, we sample a covariance matrix with appropriate scaling:
\begin{equation}
    \mathbf{\Sigma_i} \sim \text{Wishart}(\sigma\sqrt{|\kappa|}\mathbf{I}, d)
    \label{eq:rescale_vectors}
\end{equation}
where $\sigma$ is a user-specified variance scale parameter.

Finally, we sample our feature matrix $\X \in \M^{n_\text{samples} \times d}$ from wrapped normal distributions whose mean and covariance are determined by the cluster assignments in $\mathbf{c}$:
\begin{equation}
    \X = 
    \begin{pmatrix} 
        \text{---} & \mathbf{x_1} & \text{---} \\
        \text{---} & \mathbf{x_2} & \text{---}\\
         & \vdots \\
        \text{---} & \mathbf{x_n} & \text{---}
    \end{pmatrix}, \quad
    \mathbf{x_i} \sim \WN(\mathbf{m_{c_i}}, \mathbf{\Sigma_{c_i}})
\end{equation}

\paragraph{Classification Labels.}
When classifying data, we assign clusters to class labels, making sure we have at least as many clusters as labels and that each label corresponds to one or more clusters.
We establish a bijective mapping between the first $n_{\text{classes}}$ clusters and the set of class labels $\{1, 2, \ldots, n_{\text{classes}}\}$. 
For the remaining $n_{\text{clusters}} - n_{\text{classes}}$ clusters, we assign labels by sampling uniformly from $\text{DiscreteUniform}(1, n_{\text{classes}})$, thereby ensuring that the surjectivity condition of our mapping is satisfied.

This mapping is summed up in a vector $\mathbf{a} \in \{ 1, 2, \ldots, n_\text{classes} \}^{n_\text{clusters}}$.
Finally, we return:
\begin{equation}
    \y = (a_{c_1}, a_{c_2}, \ldots, a_{c_{n_\text{samples}}})
\end{equation}

\paragraph{Regression Labels.}
For regression tasks, we assign to each cluster a linear function parameterized by a slope vector $\mathbf{s}_i \in \mathbb{R}^d$ and an intercept $b_i \in \mathbb{R}$. These parameters are sampled as $\mathbf{s}_i \sim \mathcal{N}(0, 2)^d$ and $b_i \sim \mathcal{N}(0, 20)$.

Given a data point $\mathbf{x}$ assigned to cluster $c_i$, we compute its regression value using the corresponding linear function:
\begin{equation}
    y_i = \mathbf{s}_{c_i} \cdot \mathbf{x}_i + b_{c_i} + \varepsilon_i
\end{equation}
where $\varepsilon_i \sim \mathcal{N}(0, \sigma^2)$ represents additive Gaussian noise with standard deviation $\sigma$.

To facilitate the interpretation of the root mean square error (RMSE), we normalize the final regression labels to the range $[0, 1]$ via min-max scaling:
\begin{equation}
    \hat{y}_i = \frac{y_i - \min(\mathbf{y})}{\max(\mathbf{y}) - \min(\mathbf{y})}
\end{equation}

\vfill
\pagebreak
\section{Mathematical Details: Embedders}
\label{app:embedders}
An embedding is the mapping $f: U \to V$ between two metric spaces $U, V$.
We provide the \path{Manify.embedders} submodule for learning embeddings of datasets in product space manifolds.
We offer three approaches based on the user's available data:
\begin{itemize}[noitemsep,topsep=0pt]
    \item \textbf{To embed pairwise distances:} \path{Manify.embedders.coordinate_learning} trains embeddings directly from distance matrices, following \citet{gu_learning_2018}.
    \item \textbf{To embed features:} \path{Manify.embedders.vae} trains a Variational Autoencoder (VAE) with a manifold-valued latent space, as in \citet{skopek_mixed-curvature_2020}.
    \item \textbf{To embed both features and distances:} \path{Manify.embedders.siamese} trains Siamese Neural Networks, which balance reconstruction loss and distortion of pairwise distances in the latent space, as in \citet{corso_neural_2021}.
\end{itemize}

\subsection{Coordinate Learning}
We follow the coordinate learning formulation in \citet{gu_learning_2018}, the objective of which is: given a matrix $\D \in \R{n_\text{points} \times n_\text{points}}$, find an embedding $\X \in \P^{n_\text{points} \times d}$ such that the pairwise geodesic distances between the points in $\X$ closely resemble $\D$.

\paragraph{Objectives.}
To this end, we provide two fidelity measures in ---\textit{average distortion} $\Davg$, and, for graphs, \textit{mean average precision} $\mAP$:
\begin{align}
    \Davg(\X, \D) &= \frac{\sum_{i,j} |\delta_\P(\x_i, \x_j) - d_{i,j}|}{n_\text{points}^2}\\
    \mAP(\X, \G) &= \frac{1}{|V|} \sum_{a \in \V} \frac{1}{\text{deg}(a)} \sum_{i=1}^{|\mathcal{N}_a|} \frac{
        | \mathcal{N}_a \cap R_{a, b_i}|
    }{
        |R_{a,b_{i}}|
    },
\end{align}
where $R_{a,b_{i}}$ is the smallest neighborhood around node $a$ that contains its neighbor $b_i$.
Intuitively, $\Davg$ evaluates how closely pairwise distances in the embeddings match $\D$, whereas $\operatorname{mAP}$ evaluates the extent to which neighborhoods of various radii match the structure of the graph.

However, $\Davg$ or $\mAP$ are typically not optimized directly in the literature.
We follow \citet{gu_learning_2018} in substituting the auxiliary loss function
\begin{equation}
    \Ldistortion(\X, \D) = \sum_{1 \leq i \leq j \leq n} \left|
        \left(
            \frac{
                \delta_\P(\x_i, \x_j)
            }{
                D_{i,j}
            }
        \right)^2 - 1
    \right|
\end{equation}

\paragraph{Training Algorithm.}
Since we are trying to learn $\X \in \P^{\npoints \times d}$ and $\P \subset \R{d+1}$, we need some way to optimize the position of all points in $\X$ while respecting the constraint that keeps $\X \in \P$.
Typical gradient descent methods, which rely on the vector space structure of the underlying parameters, are inadequate for this.
Instead, we use Riemannian optimization (as implemented in \Geoopt\ \citep{kochurov_geoopt_2020}): in a nutshell, the gradient of the loss with respect to $\x \in \P$, $\nabla\Ldistortion$ is projected and then applied to manifold parameters using the exponential map from the tangent plane at the existing parameter value $\x$.
Here we describe how this looks for $\E{}, \S{},$ and $\H{}$:
\begin{align}
    \v = \v_\E{} &= \nabla\Ldistortion(\x, \D)\\
    \v_\S{} &= \v - \langle \v, \x \rangle_\S{} \x\\
    \v_\H{} &= (\v + \v \langle \v, \x \rangle_\H{} \x)\mathbf{J}, \text{ where }
    J_{i,j} = \begin{cases}
        -1 & \text{ if } i = j = 1\\
        1 & \text{ if } i = j \neq 1\\
        0 & \text{ otherwise}
    \end{cases}\\
    \x' &= \expmap{\x}{\alpha\v_\M},
\end{align}
where $\alpha$ is the learning rate hyperparameter.
For $\P$, these updates can be carried out separately by factorizing $\v$ and $\x$ and applying each update in its respective component.

The training algorithm itself is quite simple: We initialize $\X$ by sampling $\npoints$ points at random from $\WN(0, \I / d)$, and then update $\X$ via $\nepochs$ Riemannian optimization steps.
Optionally, the user may treat the curvatures of each of the component manifolds $\M_i \in \P$ as learnable parameters (by setting \texttt{scale\_factor\_learning\_rate} to a nonzero value).
Due to the extreme gradients in the early training steps, we include a burn-in period during which the learning rate is lower for the first several epochs.


\paragraph{Non-Transductive Training.}
We provide an optional non-transductive training mode which is absent from \citet{gu_learning_2018} and is instead inspired by \citet{chlenski_productdt_2024}.
In the non-transductive training mode, the user specifies the indices of the points that will constitute their test set, and the gradients from the test set to the training set are masked out to prevent leakage.
The test set coordinates are still optimized with respect to the entire training and test set.

Inside the training loop, we calculate train-train ($\mathcal{L}_\text{train-train}$), test-test ($\mathcal{L}_\text{test-test}$), and test-train ($\mathcal{L}_\text{train-test}$) loss, respectively.
The total loss is, straightforwardly,
\begin{equation}
    \mathcal{L}_\text{total} = \mathcal{L}_\text{train-train} + \mathcal{L}_{test-test} + \mathcal{L}_{train-test}.
\end{equation}
However, by detaching the subset of $\X$ from the Pytorch computational graph before $\mathcal{L}_{train-test}$ is back-propagated, we ensure that the points in the training set are not influenced by the test set.

\subsection{Product Space Variational Autoencoders}
We implemented the Product Space Variational Autoencoder (VAE) algorithm from \citet{skopek_mixed-curvature_2020} in \path{Manify.embedders.vae}. 

\paragraph{Setup.} To compute any VAE, we need to choose a probability distribution $p$ as a prior and a corresponding posterior distribution in the variational family $q$. 
For our prior, we use $\WN(\origin, \I)$.
Both $p$ and $q$ need to have a differentiable Kullback-Leibler (KL) divergence \citep{kingma2014vae}. 
However, the wrapped normal distribution $\WN$ does not have a closed form for the KL-divergence; therefore, we estimate the KL divergence via Monte Carlo sampling as the sample mean of the difference in log-likelihoods:
\begin{equation}
D_{\text{KL}}(q \parallel p) \approx 
\frac{1}{L} \sum_{l=1}^{L} \left( \log q(\mathbf{z^{(l)}}) - \log p(\mathbf{z^{(l)}}) \right)
\label{eq:kl_divergence}
\end{equation}
where $\mathbf{z^{(l)}} \sim q$ for all $l = 1, \ldots, L$; $\log q(\cdot)$ and $\log p(\cdot)$ denote the log-likelihood of $\WN$, as defined in Equation~\ref{eq:wn_final_log_likelihood}.

\paragraph{Implementation.}
We implemented the Product Space VAE using PyTorch in the \texttt{ProductSpaceVAE} class. The implementation follows the standard VAE architecture with encoder and decoder networks, but operates in mixed-curvature product manifold spaces. 

We assume that $\operatorname{Encoder}(\cdot)$ is any Pytorch module that maps inputs from the original input feature space $\R{D}$ to $\R{d} \times \R{d}$, and $\operatorname{Decoder}(\cdot)$ is any Pytorch module that maps inputs from $\R{d}$ to $\R{D}$.

The model encodes input data to parameters $\mathbf{z_{mean}}, \mathbf{z_{log\sigma}} \in \R{d}$ which it translates to wrapped normal distribution parameters as follows:
\begin{align}
    \mathbf{z_{mean}}, \mathbf{z_{log\sigma}} &= \operatorname{Encoder}(\mathbf{x})\\
    \boldmu &= \expmap{\origin}{\mathbf{z_{mean}}}\\
    \boldsigma &= \exp(\operatorname{diag}(\mathbf{z_{log\sigma}}))\\
    \mathbf{z} &\sim \WN(\boldmu, \boldsigma)
\end{align}
That is, $\mathbf{z}$ is a point sampled from the distribution $q = \WN(\boldmu, \boldsigma)$.
For the KL divergence term, we use Monte Carlo sampling as described in Equation~\ref{eq:kl_divergence}, drawing multiple samples from the posterior to estimate the difference between log-likelihoods.

To decode $\mathbf{z}$, we apply the logarithmic map and then feed it into the decoder network:
\begin{align}
    \mathbf{z}' &= \logmap{\origin}{\mathbf{z}}\\
    \mathbf{x_{reconstructed}} &= \operatorname{Decoder}(\mathbf{z}')
\end{align}

The model is trained by maximizing the Evidence Lower Bound (ELBO):
\begin{equation}
    \mathcal{L}_\text{ELBO}(\mathbf{x}) = 
    \underbrace{
        \mathbb{E}_{q(\mathbf{z}|\mathbf{x})}(\log p(\mathbf{x}|\mathbf{z}))
    }_{\mathcal{L}_\text{reconstruction}} - \beta
    \underbrace{
        D_{\text{KL}}(q(\mathbf{z}|\mathbf{x}) \parallel p(\mathbf{z}))
    }_{\mathcal{L}_\text{KL}}
\end{equation}
where $\mathcal{L}_{\text{reconstruction}}$ is computed using MSE loss between $\mathbf{x}$ and $\mathbf{x_{reconstructed}}$, and $\beta$ is a hyperparameter controlling the tradeoff between reconstruction accuracy and regularization. Our implementation supports arbitrary product manifolds, allowing for flexible combinations of hyperbolic, Euclidean, and spherical components.

\paragraph{Avoiding Leakage.}
Unlike with coordinate learning, avoiding train-test leakage with the VAE is straightforward: simply train the Encoder and Decoder on the training set and use the trained, frozen Encoder and Decoder to create final embeddings for the training and test sets.

\subsection{Siamese Neural Networks}
We generalize the Siamese Neural Network approach, as seen in works such as \citet{corso_neural_2021}, to product manifolds in \path{Manify.embedders.siamese}.

\paragraph{Setup.} The Siamese Neural Network architecture provides an embedding approach when both input features and target distances are available. Unlike the VAE approach which optimizes for reconstruction and KL-divergence, Siamese networks directly optimize for both reconstruction fidelity and the preservation of pairwise distances in the latent space.

Given input features $\mathbf{X} \in \mathbb{R}^{n_\text{points} \times D}$ and a target distance matrix $\mathbf{D} \in \mathbb{R}^{n_\text{points} \times n_\text{points}}$, our objective is to learn an encoder and decoder mapping between the feature space and our product manifold $\mathcal{P}$.

\paragraph{Objectives.} The Siamese network optimizes a weighted combination of two losses:
\begin{equation}
\mathcal{L}_\text{total} = \alpha 
\underbrace{
    \frac{1}{n_\text{points}} \sum_{i=1}^{n_\text{points}} \| \mathbf{x}_i - \hat{\mathbf{x}}_i \|^2
}_{\mathcal{L}_\text{reconstruction}} + (1-\alpha) 
\underbrace{
    \frac{1}{n_\text{points}^2} \sum_{i,j=1}^{n_\text{points}} \left| \delta_\mathcal{P}(\mathbf{z}_i, \mathbf{z}_j) - \mathbf{D}_{i,j} \right|
}_{\mathcal{L}_\text{distance}}
\end{equation}
where $\alpha \in [0, 1]$ controls the trade-off between reconstruction accuracy and distance preservation, $\mathbf{z}_i$ is the encoded representation of $\mathbf{x}_i$ in the product manifold, and $\hat{\mathbf{x}}_i$ is the reconstructed input.

\paragraph{Implementation.} Our implementation uses a similar architecture to our VAE model, with the encoder and decoder networks as described in the VAE section. The key difference is that there is no sampling involved as we directly map inputs to the latent space. The encoder outputs coordinates in the tangent space at the origin, which are then projected onto the product manifold using the exponential map. The decoder takes the logarithmic map of the latent representation and reconstructs the input.

\paragraph{Avoiding Leakage.} Like with VAEs, avoiding train-test leakage with Siamese networks is straightforward: we train the encoder and decoder on the training set only, and then use the trained, frozen encoder to create final embeddings for both the training and test sets. This ensures that information from the test set does not leak into the model parameters during training.

\vfill
\pagebreak
\section{Mathematical Details: Predictors}
\label{app:predictors}
\Manify\ offers a menagerie of predictors, supporting classification and regression (and occasionally link prediction) on product manifold-valued datasets.
We follow \sklearn\ \citep{pedregosa_scikit-learn_2011} conventions in our API design for our predictors.
In particular, we ensure that each predictor has \texttt{.fit()}, \texttt{.predict()}, \texttt{.score()}, and (for classifiers) \texttt{.predict\_proba()} methods available.

We implement the following predictors for product space manifolds:
\begin{itemize}[noitemsep,topsep=0pt]
    \item \textbf{Decision Trees and Random Forests}, as in \citet{chlenski_productdt_2024}.
    \item \textbf{Graph Convolutional Networks}, as in the $\kappa$-GCN described in \citet{bachmann_constant_2020}. 
    By ablating certain components of the $\kappa$-GCN, we can additionally create \textbf{Multi-Layer Perceptrons} and \textbf{Logistic/Linear Regression}.
    \item \textbf{Perceptrons}, as in \citet{tabaghi_linear_2021}.
    \item \textbf{Support Vector Machines}, as in \citet{tabaghi_linear_2021}.
    \item \textbf{Mixed-Curvature Stereographic Transformers}, as in \citet{cho_curve_2023}.
\end{itemize}

\subsection{Product Space Decision Trees and Random Forests}
We implemented the ProductDT algorithm from \citet{chlenski_productdt_2024}, which extends \citet{chlenski_fast_2024} to product space manifolds, in \path{Manify.predictors.decision_tree}.
These works are both non-Euclidean versions of Decision Trees \citep{breiman_classification_2017} and Random Forests \citep{breiman_random_2001}, classic machine learning algorithms.

\paragraph{Decision Boundaries in Product Spaces.} 
The core innovation of ProductDT is adapting decision boundaries to the geometry of product manifolds. While traditional decision trees partition Euclidean space using hyperplanes, ProductDT uses geodesic decision boundaries appropriate for each component manifold in the product space.

For points in a product manifold $\mathcal{P} = \mathcal{M}_1 \times \mathcal{M}_2 \times \cdots \times \mathcal{M}_k$, we define splits along two-dimensional subspaces for each component manifold. This approach preserves the computational efficiency of traditional decision trees while respecting the underlying manifold geometry.

\paragraph{Angular representation of splits.}
For each component manifold $\mathcal{M}_i$, we project points onto two-dimensional subspaces and represent potential splits using angles. Given a point $\mathbf{x}$ and a basis dimension $d$, the projection angle is computed as:
where $x_0$ and $x_d$ are the coordinates in the selected two-dimensional subspace.
The splitting criterion then becomes:
\begin{equation}
    S(\mathbf{x},d,\theta)=\begin{cases}
        1 &\text{ if } \tan^{-1}\left( \frac{x_0}{x_d} \right)\in[\theta,\theta+\pi)\\
        0 &\text{ otherwise}
    \end{cases}.
    \label{criterion}
\end{equation}

\paragraph{Geodesic midpoints for decision boundaries.}
To place decision boundaries optimally between clusters of points, we compute the geodesic midpoint between consecutive angles in the sorted list of projection angles. The midpoint calculation is specific to each manifold type:

\begin{align}
    \theta_u &= \tan^{-1}\left(\frac{u_0}{u_d}\right) \\
    \theta_v &= \tan^{-1}\left(\frac{v_0}{v_d}\right)\\
    m_{\mathbb{E}}(\theta_u, \theta_v) &= \tan^{-1}\left( \frac{2}{u_0 + v_0} \right)\\
    m_{\mathbb{S}}(\theta_u, \theta_v) &= \frac{\theta_u + \theta_v}{2}\\
    m_{\mathbb{H}}(\theta_u, \theta_v) &= \begin{cases}
        \cot^{-1}\left(V - \sqrt{V^2-1}\right) &\text{ if } \theta_u + \theta_v < \pi, \\
        \cot^{-1}\left(V + \sqrt{V^2-1}\right) &\text{ otherwise.}
    \end{cases}\\
    V &= \frac{\sin(2\theta_u - 2\theta_v)}{2\sin(\theta_u + \theta_v)\sin(\theta_v - \theta_u)}
\end{align}

These formulas ensure that decision boundaries are placed at points geodesically equidistant from the nearest points on either side, respecting the intrinsic geometry of each manifold component.

\paragraph{Training algorithm.}
The ProductDT algorithm follows the recursive structure of classical decision trees:
\begin{enumerate}[noitemsep,topsep=0pt]
    \item For each component manifold in the product space, compute projection angles for all points along each basis dimension.
    \item For each projection, sort the angles and evaluate potential splits using the geodesic midpoints between consecutive angles.
    \item Select the split that optimizes a chosen impurity criterion (e.g., Gini impurity or information gain).
    \item Recursively apply the procedure to each resulting partition until a stopping criterion is met (maximum depth, minimum samples per leaf, etc.).
\end{enumerate}

This decomposition leverages the structure of product spaces to maintain the $\mathcal{O}(1)$ decision complexity at inference time while adapting to the underlying geometry of the data.

\paragraph{Random Forests.}
Much like their Euclidean (and hyperbolic) counterparts, Product Space Random Forests are simply ensembles of bootstrapped Decision Trees.
An implementation of Product Space Random Forests is available as the \texttt{ProductSpaceRF} class in the \path{Manify.predictors.decision_tree} submodule.

\subsection{Product Space Graph Convolutional Networks}
We implemented the $\kappa$-GCN algorithm in \path{Manify.predictors.kappa_gcn} from \citet{bachmann_constant_2020}, who introduces gyrovector operation that extends existing frameworks \citep{chami_hyperbolic_2019} that are limited to negative curvature to both negative and positive values that interpolates smoothly. We start by introducing these concepts. 

\paragraph{Graph Convolutional Networks Background.}
In a typical (Euclidean) graph convolutional network (GCN), each layer takes the form
\begin{align}
    \mathbf{H}^{(0)} &= \X\\
    \mathbf{H}^{(l+1)} &= \sigma\left( \Ahat \mathbf{H}^{(l)} \W^{(l)} + \mathbf{b}^{(l)} \right),
\end{align}
where $\Ahat \in \R{\npoints \times \npoints}$ is a normalized adjacency matrix with self-connections, $\X^{(l)} \in \R{\npoints \times d}$ is a matrix of features, $\W^{(l)} \in \R{d \times e}$ is a weight matrix, $\mathbf{b}^{(l)} \in \R{e}$ is a bias term, and $\sigma$ is some nonlinearity, e.g. ReLU.

\paragraph{Graph Convolution Layers.}
\noindent \citet{bachmann_constant_2020} describes a way to adapt the typical GCN model for use with $\X \in \st{d}{\kappa}$, in terms of gyrovector operations:
\begin{align}
    \mathbf{H}^{(l+1)} &= \sigma^{\otimes_\kappa} \left( \hat{\mathbf{A}} \boxtimes_\kappa \left( \mathbf{H}^{(l)} \otimes_\kappa \mathbf{W}^{(l)} \right) \right),\\
    \sigma^{\otimes_\kappa}(\cdot) &= \expmap{\origin}{
        \sigma(\logmap{\origin}{\cdot})
    }
\label{layers}
\end{align}
Note that this paper does not include a bias term, although it is reasonable to extend the definition of a GCN layer to include one:
\begin{equation}
    \mathbf{H}^{(l+1)} = 
    \sigma^{\otimes_\kappa} \left( 
        \hat{\mathbf{A}} \boxtimes_\kappa \left(
            \mathbf{H}^{(l)} \otimes_\kappa \mathbf{W}^{(l)} 
        \right) \oplus \mathbf{b}
    \right),
\end{equation}
where $\mathbf{b} \in \st{d}{\kappa}$ is a bias vector.
Also note that, in order for each $\mathbf{H}^{(i)}$ to stay on the same manifold, $\W^{(i)}$ must be a square matrix; however, it is possible to relax this assumption and specify separate dimensionalities and curvatures for each layer instead.

\paragraph{Stereographic Logits.}
For classification, we need to define a $\kappa$-sterographic equivalent of a logit layer:
\begin{equation}
    \mathbf{H}^{(L)} = \text{softmax} \left( \hat{\mathbf{A}} \operatorname{logits}_{\st{d}{\kappa}} \left( \mathbf{H}^{(L-1)}, \mathbf{W}^{(L-1)} \right) \right).
\label{final_layers}
\end{equation}

In order to implement logits in $\st{d}{\kappa}$, we first consider the interpretation that Euclidean logits are signed distances from a hyperplane.
This follows naturally from the linear form $\mathbf{w}_i^\top\mathbf{x} + b_i$ used in traditional classification, where $\mathbf{w}_i$ is a column vector of the final weight matrix and $b_i$ is its corresponding bias. 
The magnitude directly corresponds to the point's distance from the decision boundary (the hyperplane defined by $\mathbf{w}_i^\top\mathbf{x} + b_i = 0$), and the sign indicates which side of the hyperplane the point lies on—thus encoding both the model's decision and confidence in a single value.

\citet{bachmann_constant_2020} and \citet{ganea_hyperbolic_2018} use this perspective to redefine logits in non-Euclidean spaces by substituting the appropriate distance function.
In $\kappa$-GCN, this becomes:
\begin{align}
    \mathbb{P}(y = k\ |\ \mathbf{x}) &= \operatorname{Softmax}\left( \operatorname{logits}_\M(\x, k) \right)\\
    \operatorname{logits}_\M(\x, k) &= \frac{\|\mathbf{a_k}\|_{\mathbf{p_k}}}{\sqrt{K}} \sin_K^{-1}\left(
        \frac{2\sqrt{|\kappa|}\langle \mathbf{z_k}, \mathbf{a_k} \rangle}{(1 + \kappa\|\mathbf{z_k}\|^2)\|\mathbf{a_k}\|}
    \right),
\end{align}
where $\mathbf{a_k}$ is a column vector of the final weight matrix corresponding to class $k$, $\mathbf{p_k} \in \st{d}{\kappa}$ is a bias term, and $\mathbf{z_k} = -\mathbf{p_k} \oplus \x$.

Although it is not explicit in \citet{bachmann_constant_2020} how the logits aggregate across different product manifolds, we follow \citet{cho_curve_2023} and later \citet{chlenski_productdt_2024} in aggregating logits as the $\ell_2$-norm of component manifold logits, multiplied by the sign of the sum of the component inner products:
\begin{equation}
    \operatorname{logits}_\P(\x, k) = \sqrt{\sum_{\M \in \P} \operatorname{logits}_\M(\x_\M, k)} \cdot \sum_{\M \in \P} \langle \x_\M, \mathbf{a_k}_\M \rangle.
\end{equation}
This is consistent with the interpretation of logits as signed distances to a hyperplane, generalized to product space manifolds.

\paragraph{Generating Adjacency Matrices.}
Following \citet{bachmann_constant_2020}, we offer the convenience function \path{Manify.predictors.kappa_gcn.get_A_hat} to preprocess any square adjacency matrix $\A \in \R{d \times d}$ into a normalized, symmetric adjacency matrix with self-connections appropriate for use with a GCN as follows:
\begin{align}
    \mathbf{A}' &= \mathbf{A} + \mathbf{A}^\top\\
    \tilde{\mathbf{A}} &= \mathbf{A}'+ \mathbf{I}\\
    \tilde{D}_{ii} &= \sum_k \tilde A_{ik}\\
    \hat{\mathbf{A}} &= \tilde{\mathbf{D}}^{-1/2} \tilde{\mathbf{A}} \tilde{\mathbf{D}}^{-1/2},
\end{align}

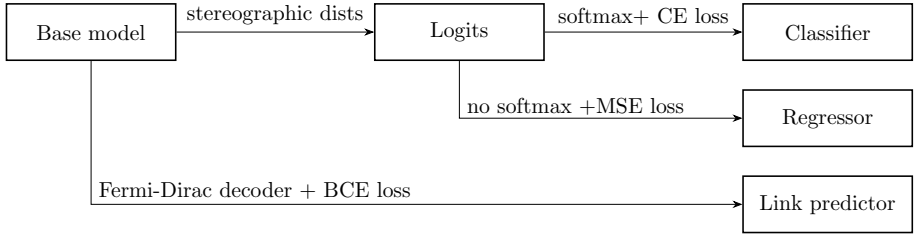
\begin{figure}[t]
    \centering
    \scalebox{.75}{\begin{tikzpicture}[
    box/.style={draw, minimum width=3cm, minimum height=1cm, thick},
    >=Stealth,
    node distance=0.5cm and 3.5cm
]

\node[box] (base) {Base model};
\node[box, right=of base] (logits) {Logits};
\node[box, right=of logits] (classifier) {Classifier};
\node[box, below=of classifier] (regressor) {Regressor};
\node[box, below=of regressor] (link) {Link predictor};

\draw[->] (base) -- node[above] {stereographic dists} (logits);
\draw[->] (logits) -- node[above] {softmax+ CE loss} (classifier);

\draw[->] (logits) -- node[right, pos=0.8] {no softmax +MSE loss} (logits |- regressor.west) -- (regressor);
\draw[->] (base) -- node[right, pos=0.9] {Fermi-Dirac decoder + BCE loss} (base |- link.west) -- (link);
\end{tikzpicture}}
    \caption{The relationships between the classification, regression, and link prediction modes of the $\kappa$-GCN: both the classifier and the regressor depend on stereographic hyperplane distances (logits); link predictions go through the Fermi-Dirac decoder instead.}
    \label{fig:task_relationships}
\end{figure}

\paragraph{Predictions with $\kappa$-GCN.}
The stereographic logits described above can be turned into classification targets through a standard softmax function.
This is the only task for which $\kappa$-GCN was described, but there are precedents in the literature for using such models for regression and link prediction tasks.
We describe these variants below, and summarize the relationships between these three tasks in Figure~\ref{fig:task_relationships}.

For regression problems, we follow \citet{liu_hyperbolic_2019} in setting the output dimension of our $\kappa$-GCN to 1, omitting the final Softmax operation, and training with a mean squared error loss function.

Finally, for link prediction, we follow \citet{chami_hyperbolic_2019} in adopting the standard choice of applying the Fermi-Dirac decoder \citep{krioukov_hyperbolic_2010, nickel_poincare_2017} to predict edges:
\begin{equation}
    \label{eq:fermi-dirac}
    \mathbb{P}((i,j) \in \mathcal{E} | \mathbf{x_i}, \mathbf{x_j}) = \left(
        \exp\left(
            \frac{\delta_\M(\mathbf{x_i}, \mathbf{x_j})^2 - r}{t}
        \right) + 1
    \right)^{-1},
\end{equation}
where the embeddings for points $i$ and $j$, $\mathbf{x_i}$ and $\mathbf{x_j}$, may be updated by $\kappa$-GCN layers.

\paragraph{Deriving MLPs and (Multinomial Logistic) Regression from $\kappa$-GCN.}
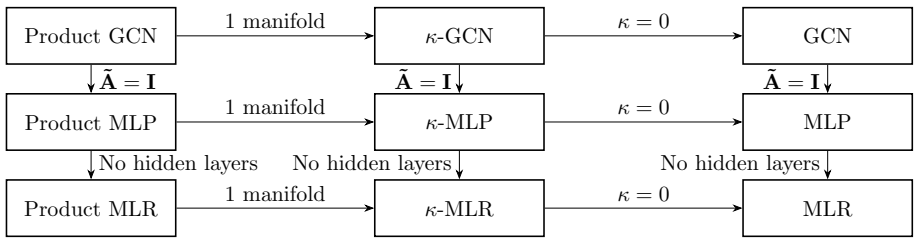
\begin{figure}[t]
    \centering
    \scalebox{.75}{\begin{tikzpicture}[
    box/.style={draw, minimum width=3cm, minimum height=1cm, thick},
    >=Stealth,
    node distance=0.5cm and 3.5cm
]

\node[box] (product_gcn) {Product GCN};
\node[box, right=of product_gcn] (kappa_gcn) {$\kappa$-GCN};
\node[box, right=of kappa_gcn] (gcn) {GCN};

\node[box, below=of product_gcn] (product_mlp) {Product MLP};
\node[box, right=of product_mlp] (kappa_mlp) {$\kappa$-MLP};
\node[box, right=of kappa_mlp] (mlp) {MLP};

\node[box, below=of product_mlp] (product_mlr) {Product MLR};
\node[box, right=of product_mlr] (kappa_mlr) {$\kappa$-MLR};
\node[box, right=of kappa_mlr] (mlr) {MLR};

\draw[->] (product_gcn) -- node[above] {1 manifold} (kappa_gcn);
\draw[->] (kappa_gcn) -- node[above] {$\kappa = 0$} (gcn);

\draw[->] (product_mlp) -- node[above] {1 manifold} (kappa_mlp);
\draw[->] (kappa_mlp) -- node[above] {$\kappa = 0$} (mlp);

\draw[->] (product_mlr) -- node[above] {1 manifold} (kappa_mlr);
\draw[->] (kappa_mlr) -- node[above] {$\kappa = 0$} (mlr);

\draw[->] (product_gcn) -- node[right] {$\mathbf{\tilde{A}} = \mathbf{I}$} (product_mlp);
\draw[->] (kappa_gcn) -- node[left] {$\mathbf{\tilde{A}} = \mathbf{I}$} (kappa_mlp);
\draw[->] (gcn) -- node[left] {$\mathbf{\tilde{A}} = \mathbf{I}$} (mlp);

\draw[->] (product_mlp) -- node[right] {No hidden layers} (product_mlr);
\draw[->] (kappa_mlp) -- node[left] {No hidden layers} (kappa_mlr);
\draw[->] (mlp) -- node[left] {No hidden layers} (mlr);

\end{tikzpicture}}
    \caption{We diagram the relationship between 9 different models, demonstrating how they are all special cases of $\kappa$-GCN. In particular, the horizontal axis tracks the relationship $\E{} \subset \M{} \subset \P{}$, whereas the vertical axis tracks the relationship $\operatorname{MLR} \subset \operatorname{MLP} \subset \operatorname{GCN}$. We use ``MLR'' to refer generically to all one-layer models.}
    \label{fig:model_relationships}
\end{figure}
We note that $\kappa$-GCNs contain 8 other models as special cases, which we diagram in Figure~\ref{fig:model_relationships}.
In particular, any GCN can be transformed into an MLP by setting $\Ahat = \I$ (effectively not passing messages along the graph edges), and any MLP can further be turned into a (multinomial logistic) regressor be removing all of its hidden layers.
This has been noted in other work, such as \citet{wu_simplifying_2019}.

\subsection{Product Space Perceptrons}
\label{app:perceptron}
Product Space Perceptrons are currently offered as an experimental feature.
Support is limited, and we have not thoroughly validated our implementation.
Interested readers should refer to \citet{tabaghi_linear_2021} for full details.

A linear classifier on $\P$ is defined as
\begin{equation}
    \operatorname{LC}(\x, \w) = \operatorname{sign}\left( 
        \langle \w_\E{}, \x_\E{} \rangle + \alpha_\S{} \sin^{-1}\left( \langle \w_\S{}, \x_\S{} \rangle \right) + \alpha_\H{} \sinh^{-1}\left( \langle \w_\H{}, \x_\H{} \rangle_\H{} \right) + b,
    \right)
\end{equation}
where $\X_\M$ ($\x_\M$) denotes the restriction of $\X \in \P$ ($\x \in \P$) to one of its component manifolds, $\alpha_\S{}$ and $\alpha_\H{}$ are weights, and $b \in \R{}$ is a bias term.
We further require that $\| \w_\E{} \| = \alpha_\E{}$, $\| \w_\S{} \| = \sqrt{\kappa_\S{}}$, and $\| \w_\H{} \| = \sqrt{-\kappa_\H{}}$.

The Product Space Perceptron formulation relies on a kernel function.
Letting $R = \max_{\x_n \in \X} \|\x_{\H{},n}\|_2$, that is the maximum \textit{Euclidean} norm of any hyperbolic component of $\X$, we define the kernel function as
\begin{align}
    K(\X, \mathbf{x_n}) &= 
    1 + 
    \X_{\E{}}^\top\x_{\E{},n} + 
    \alpha_\S{} \sin^{-1}\left( \kappa_\S{} \X_{\S{}}^\top \x_{\S{},n} \right) + \alpha_\H{} \sin^{-1}\left( R^{-2}\X_{\H{}}^\top \x_{\H{},n} \right),\\
    \mathbf{K} &= \begin{pmatrix}
        \text{---} & K(\X, \mathbf{x_1}) & \text{---}\\
        \text{---} & K(\X, \mathbf{x_2}) & \text{---}\\
        & \vdots \\
        \text{---} & K(\X, \mathbf{x_{\npoints}}) & \text{---}\\
    \end{pmatrix},
\end{align}
where $\X \in \P^{\npoints \times d}$ is a set of points on the manifold.

Using this kernel, which we implement in \path{Manify.predictors.kernel}, we can train a classic kernel perceptron to work in the product space. 

\subsection{Product Space Support Vector Machines}
Product Space Support Vector Machines (SVMs) are currently offered as an experimental feature.
Support is limited, and we have not thoroughly validated our implementation.
Interested readers should refer to \citet{tabaghi_linear_2021} for full details.

The Product Space SVM extends the kernel approach of Section~\ref{app:perceptron}by finding a maximum-margin classifier in the product space. 
The optimization problem is formulated as:
\begin{align}
    \text{maximize } & \varepsilon - \sum_{i=1}^{\npoints} \xi_i,\\
    \text{subject to } & y_i \sum_{j=1}^{\npoints} \beta_j K(\x_i, \x_j) \geq \varepsilon - \xi_i \text{ for all } i \in \{1,\ldots,\npoints\},\\
    \text{where } & \varepsilon > 0 \text{ and } \xi_i \geq 0
\end{align}
Additionally, we constrain the weight vector $\beta$ to satisfy:
\begin{align}
    \beta &\in \text{convhull}(\mathcal{A}_\E{}) \cap \text{convhull}(\mathcal{A}_\S{}) \cap \widetilde{\mathcal{A}_\H{}}, \text{ where} \\[0.5em]
    \mathcal{A}_\E{} &= \left\{ \beta \in \mathbb{R}^{\npoints} : \beta^\top \kappa_{\E{}} \beta = \alpha_{\E{}}^2 \right\}, \\[0.5em]
    \mathcal{A}_\S{} &= \left\{ \beta \in \mathbb{R}^{\npoints} : \beta^\top \kappa_{\S{}} \beta = \frac{\pi}{2} \right\}, \\[0.5em]
    \mathcal{A}_\H{} &= \left\{ \beta \in \mathbb{R}^{\npoints} : \beta^\top \kappa_{\H{}} \beta = \sinh^{-1}(-R^2\kappa_\H{}) \right\}.
\end{align}
For practical optimization, we follow \citet{tabaghi_linear_2021} in replacing the Euclidean and spherical constraint sets with their convex hulls, and relax the non-convex hyperbolic constraints through a decomposition approach. 

\subsection{Mixed-Curvature Stereographic Transformers}
We implemented the Fully Product-Stereographic Transformer (FPS-T) algorithm in \path{Manify.predictors.nn.layers} following the approach proposed by \citet{cho_curve_2023}. The implementation is still preliminary and has not been rigorously validated. FPS-T extends the attention mechanism of standard transformers to operate on mixed-curvature spaces using the stereographic model. Each attention head operates on a stereographic space with learnable curvature that can represent Euclidean, hyperbolic, and spherical geometries. 

Given a sequence of $n$ product-stereographic embeddings, the stereographic multi-head attention mechanism first computes queries ($\mathbf{Q}$), keys ($\mathbf{K}$), and values ($\mathbf{V}$) by mapping each stereographic embedding to the tangent space of the values. This key insight enables the geometric operations necessary to compute attention scores in stereographic space. Specifically, the attention scores between $\mathit{i}$-th query $\mathbf{Q}_i$ and $\mathit{j}$-th key $\mathbf{K}_i$ are computed by parallel transporting the query and key vectors to the origin and taking their Riemannian inner product:
\begin{align}
    \alpha_{ij} &= \langle \text{PT}_{V_i \to 0}(\mathbf{Q}_i), \text{PT}_{V_j \to 0}(\mathbf{K}_j) \rangle
\end{align}
where $\text{PT}_{V_i \to 0}$ denotes parallel transport from point $V_i$ to the origin, which preserves the geometric structure during the transport operation. The values are then aggregated using the Einstein midpoint, which is the geometric generalization of weighted averaging in curved spaces:
\begin{align}
    \text{Agg}_\kappa(\mathbf{V}, \boldsymbol{\alpha})_i &= \frac{1}{2} \oplus_\kappa \left[ \sum_{j=1}^n \alpha_{ij} \frac{\lambda_\kappa^{\mathbf{V_j}}}{\sum_{k=1}^n \alpha_{ik}(\lambda_\kappa^{\mathbf{V_k}} - 1)} \mathbf{V}_j \right]
\end{align}
where $\oplus_\kappa$ is the Möbius addition operation, and $\lambda_\kappa^x = \frac{4}{1 + \kappa\|x\|^2}$ is the conformal factor that accounts for the curvature at point $x$. The final multi-head attention output concatenates results from all $H$ attention heads:
\begin{equation}
    \text{MHA}_{\otimes^\kappa}(\mathbf{X}) = \big\Vert_{h=1}^H \text{Aggregate}_{\kappa_h}(\mathbf{V}^h, \mathbf{\alpha}^h)
\end{equation}
The model learns curvatures end-to-end through gradient descent, automatically adapting the geometry to match the input graph structure.

\vfill
\pagebreak
\section{Mathematical Details: Curvature Estimation}
\label{app:curvature}
Given a set of distances, one might wish to ask what the curvature of the underlying space is.
We implement two popular approaches to answering this question:
\begin{itemize}[noitemsep,topsep=0pt]
    \item \textbf{Sectional curvature} in \path{Manify.curvature_estimation.sectional_curvature}, to measure the (positive or negative) curvature for specific nodes in a graph, and
    \item \textbf{Delta-hyperbolicity}, in \path{Manify.curvature_estimation.delta_hyperbolicity}, to measure how negatively-curved an entire space is.
    \item{Greedy heuristics}, in \path{Manify.greedy_method}, iteratively select submanifolds which minimize some objective function at each step.
\end{itemize}

\subsection{Sectional curvature}
We follow \citet{gu_learning_2018} in our definition of sectional curvature, which we implement in \path{Manify.curvature_estimation.sectional_curvature}.
Given a geodesic triangle between $\mathbf{a}, \mathbf{b}, \mathbf{c} \in \M$, the curvature (derived from the parallelogram law in Euclidean geometry) is
\begin{equation}
    \xi_\M(\mathbf{a}, \mathbf{b}, \mathbf{c}) = 
    \delta_\M(\mathbf{a}, \mathbf{m})^2 + \frac{\delta_\M(\mathbf{b}, \mathbf{c})^2}{4} - \frac{\delta_\M(\mathbf{a}, \mathbf{b})^2 + \delta_\M(\mathbf{a}, \mathbf{c})^2}{2},
    \label{eq:sectional_gu}
\end{equation}
where $\mathbf{m}$ is the midpoint of the geodesic connecting $\mathbf{b}$ and $\mathbf{c}$.

Of course, if we were able to take geodesic midpoints between points, we would already know the curvature of our space.
Therefore, it is more interesting to consider that we only know some distances between select points.
This can be described using distances in graphs.
Given nodes $a, b, c, m \in \V$ such that $(b, c) \in \edges$, we can compute the graph equivalent of sectional curvature:
\begin{align}
    \xi_\G(m;b,c;a) &= 
    \frac{1}{2\delta_G(a,m)}\left(
        \delta_\G(a, m)^2 + \frac{\delta_\G(b, c)^2}{4} - \frac{\delta_\G(a, b)^2 + \delta_\G(a, c)^2}{2}
    \right)\\
    \xi_\G(m;b,c) &= \frac{1}{|\V|-1} \sum_{a \neq m} \xi_\G(m;b,c;a).
\end{align}
Note that $\xi_\G(m;b,c)$ is just an average over $\xi_\G(m;b,c;a)$ for all $a \in \V$.

\subsection{$\delta$-hyperbolicity}
While sectional curvature describes local curvature, $\delta$-hyperbolicity is a notion of global negative curvature.
In particular, $\delta$-hyperbolicity measures the extent to which distances in a space deviate from distances in an idealized tree spanning those same points. 
For simplicity, we will continue to use the formulation on graphs, noting that any distance matrix can be thought of as a dense graph.

First, we define the Gromov product of $x, y, z \in \V$:
\begin{equation}
    (y, z)_x = \frac{1}{2} \left(\delta_\G(x, y) + \delta_\G(x, z) - \delta_\G(y, z) \right)
    \label{delta_between_yz}.
\end{equation}
This gives rise to a definition of $\delta$ as the \textit{minimal} value such that
\begin{equation}
    \forall x,y,z,w \in \V,\quad (x, z)_w \geq \min \left( (x, y)_w, (y, z)_w \right) - \delta
    \label{hyperbolcity_simplified}.
\end{equation}

In practice, a typical simplification is to to fix $w = w_0$, reducing the complexity of the $\delta$-hyperbolicity computation from $\mathcal{O}(n^4)$ to $\mathcal{O}(n^3)$.
In practice, delta is computed as
\begin{align}
    \delta &= \max\left\{ \max_k\min\{G_{i,k}, G_{k,j}\} - \mathbf{G}\right\}, \text{ where}\\
    \mathbf{G} &= \begin{pmatrix}
        (x_0, x_0)_{w_0} & (x_0, x_1)_{w_0} & \ldots & (x_0, x_n)_{w_0}\\
        (x_1, x_0)_{w_0} & (x_1, x_1)_{w_0} & \ldots & (x_1, x_n)_{w_0}\\
        \vdots           &  \vdots          & \ddots & \vdots \\
        (x_n, x_0)_{w_0} & (x_n, x_1)_{w_0} & \ldots & (x_n, x_n)_{w_0}\\
    \end{pmatrix}.
\end{align}

For performance reasons, we include a fast vectorized implementation of the $\delta$-hyperbolicity calculation, as well as a sampling-based calculation in which we consider some number $\nsamples \ll \npoints^4$ of 4-tuples of points.

\subsection{Greedy signature selection}
We follow \citet{tabaghi_linear_2021} in our formulation of greedy signature selection.
The core idea of this algorithm is that, given some (black-box) pipeline which returns a scalar, we can iteratively build up a near-optimal product manifold $\P$ from a set of candidate submanifolds (by default, $\{\H{2}_{-1},\ \E{2},\ \S{2}_1\}$) for a set number of iterations.
At each iteration, whichever submanifold achieves the lowest loss in the pipeline gets added to the optimal signature, and the process repeats; if none of the submanifolds beat the baseline manifold, the algorithm terminates.

Although users are encouraged to write their own pipelines for greedy signature selection, we provide two pipelines as a starting point in \path{Manify.curvature_estimation._pipelines}:
\begin{itemize}[noitemsep,topsep=0pt]
    \item \textbf{The distortion-minimization pipeline} is implemented in \path{Manify.curvature_estimation._pipelines.distortion_pipeline}, and allows users to find a signature that minimizes distortion for a given matrix of distances $\mathbf{D}$ when embedded using the Coordinate Learning embedder. 
    \item \textbf{The predictor score-maximization pipeline} is implemented in \path{Manify.curvature_estimation._pipelines.predictor_pipeline}, and handles embedding and classifying (or regressing on) graph-based data. Given a matrix of pairwise distances $\mathbf{D}$ and labels $\mathbf{y}$, this pipeline generates embeddings, trains a classifier on the embeddings, and optimizes the accuracy (or MSE, for regression) of the predictor.
\end{itemize}

\vfill
\pagebreak
\section{Mathematical Details: Clustering and Optimizers}
\label{app:clustering}

This appendix provides a brief mathematical summary of the algorithms contributed by \citet{yuan_riemannian_2025} and integrated into the \Manify library:
\begin{itemize}[noitemsep,topsep=0pt]
    \item \textbf{Riemannian Fuzzy $K$-Means} is implemented in \path{Manify.clustering.fuzzy_kmeans}, and provides an efficient variant of $K$-means clustering designed to work with product manifolds, and
    \item \textbf{Riemannian Adan} is implemented in \path{Manify.optimizers.radan}, and provides a Riemannian variant of the Adan algorithm. It is used to speed up the Riemannian Fuzzy $K$-Means algorithm.
\end{itemize}

\subsection{Riemannian Fuzzy $K$-Means (RFK)}

The Riemannian Fuzzy $K$-Means (RFK) algorithm is an efficient clustering method for data residing on a Riemannian manifold $\mathcal{M}$. 
It overcomes the significant computational hurdles associated with a naive extension of the standard Fuzzy $K$-Means (also known as Fuzzy $c$-Means) algorithm.

\subsubsection{The Naive Extension and its Limitations}

Unlike standard $K$-means clustering, Fuzzy $K$-means allows for ``fuzzy,'' i.e. partial, cluster assignments for each point.
The objective function for standard Fuzzy $K$-Means in Euclidean space is:
\begin{equation}
    J_m = \sum_{i=1}^{N} \sum_{j=1}^{C} u_{ij}^m \|x_i - c_j\|^2,
\end{equation}
where $\{x_i\}_{i=1}^N$ are the data points, $\{c_j\}_{j=1}^C$ are the cluster centers, $u_{ij} \in [0, 1]$ is the fuzzy membership of point $x_i$ in cluster $c_j$, and the exponent $m > 1$ controls the ``fuzziness'' of class assignments.

The naive extension to a Riemannian manifold (Naive Fuzzy $K$-Means, or NFK) replaces the Euclidean distance with the geodesic distance $\delta_\M(\cdot, \cdot)$ on the manifold $\M$:
\begin{equation}
    J_m(\mathbf{U}, \mathbf{C}) = \sum_{i=1}^{N} \sum_{j=1}^{C} u_{ij}^m \delta_\M(\mathbf{x_i}, \mathbf{c_j})^2, \quad \text{s.t.} \quad \mathbf{c}_j \in \mathcal{M},\ \sum_{j=1}^{C} u_{ij} = 1.
\end{equation}
While the update for the membership matrix $\mathbf{U} = [u_{ij}]$ retains a closed-form solution, the update for the cluster centers $\mathbf{c_j}$ becomes a Fréchet mean problem, which must be solved with an iterative Riemannian optimization routine:
\begin{equation}
    \mathbf{c_j} = \arg\min_{\mathbf{c} \in \M} \sum_{i=1}^{N} u_{ij}^m \delta_\M(\mathbf{x_i}, \mathbf{c})^2.
\end{equation}
This approach is computationally expensive because it requires running a full Riemannian optimization for each cluster center in every iteration of the main algorithm. 
This leads to a time complexity of $\mathcal{O}(MN)$, where $M$ is the number of main clustering iterations and $N$ is the cost of a single Riemannian optimization.

RFK reformulates the objective function to avoid repeated optimization by expressing the optimal membership $u_{ij}$ as a direct function of the cluster centers $\{\mathbf{c_j}\}$:
\begin{equation}
    u_{ij}(\mathbf{C}) = \left( 
        \sum_{k=1}^{C} \left( \frac{\delta_\M(\mathbf{x_i}, \mathbf{c_j})}{\delta_\M(\mathbf{x_i}, \mathbf{c_k})} 
        \right)^{\frac{2}{m-1}} \right)^{-1}.
\end{equation}
By substituting this expression back into the objective function, the problem is transformed into a single optimization problem that depends only on the cluster centers $\mathbf{C}$:
\begin{align}
    J_m(\mathbf{C}) &= \sum_{i=1}^{N} \sum_{j=1}^{C} \left( 
        \left( 
            \sum_{k=1}^{C} \left( 
                \frac{\delta_\M(\mathbf{x_i}, \mathbf{c_j})}{\delta_\M(\mathbf{x_i}, \mathbf{c_k})} 
            \right)^{\frac{2}{m-1}} 
        \right)^{-1} 
    \right)^m \delta_\M(\mathbf{x_i}, \mathbf{c_j})^2 \\
    \label{eq:rfk_final_obj}
    &= \sum_{i=1}^{N} \left( \sum_{j=1}^{C} \delta_\M(\mathbf{x_i}, \mathbf{c_j})^{\frac{2}{1-m}} \right)^{1-m}.
\end{align}
The resulting objective function in Equation \ref{eq:rfk_final_obj} is differentiable with respect to the cluster centers $\mathbf{c_j}$. 
It can therefore be minimized using a single, continuous Riemannian gradient-based optimization run, reducing the time complexity from $\mathcal{O}(MN)$ to $\mathcal{O}(N)$.

\subsection{The Radan Optimizer}

To efficiently solve the Riemannian optimization problem posed by the RFK algorithm, \citet{yuan_riemannian_2025} introduce the Radan optimizer. 
Radan is a Riemannian adaptation of the Adaptive Nesterov Momentum (Adan) algorithm \citep{xie_adan_2024}, an effective optimizer in Euclidean space.

Radan extends Adan to operate on Riemannian manifolds in the standard Riemannian optimization fashion: by integrating the exponential map and parallel transport.
The update rules for Radan at iteration $t$ are as follows:
\begin{align}
    \mathbf{m}_t &= \beta_{1,t} \varphi(\mathbf{m}_{t-1}) + (1 - \beta_{1,t}) \mathbf{g}_t \\
    \mathbf{v}_t &= \beta_{2,t} \varphi(\mathbf{v}_{t-1}) + (1 - \beta_{2,t}) (\mathbf{g}_t - \varphi(\mathbf{g}_{t-1})) \\
    \mathbf{z}_t &= \mathbf{g}_t + \beta_{2,t} (\mathbf{g}_t - \varphi(\mathbf{g}_{t-1})) \\
    n_t &= \beta_3 n_{t-1} + (1 - \beta_3) \|\mathbf{z}_t\|^2 \\
    \mathbf{u}_t &= \mathbf{m}_t + \beta_{2,t} \mathbf{v}_t \\
    \alpha_t &= \frac{\eta_t}{\sqrt{n_t} + \epsilon} \\
    \mathbf{x}_{t+1} &= \exp_{\mathbf{x}_t} (-\alpha_t \mathbf{u}_t),
\end{align}
where:
\begin{itemize}[noitemsep,topsep=0pt]
    \item $\mathbf{x}_t$ is the current point (parameter) on the manifold.
    \item $\mathbf{g}_t$ is the Riemannian gradient of the loss function at $\mathbf{x}_t$.
    \item $\varphi(\cdot)$ denotes parallel transport from tangent space $T_{\mathbf{x}_{t-1}}\mathcal{M}$ to $T_{\mathbf{x}_t}\mathcal{M}$.
    \item $\mathbf{m}_t$ is the first-order moment (momentum) of the gradients.
    \item $\mathbf{v}_t$ is the first-order moment of the \emph{gradient difference}, accelerating convergence.
    \item $n_t$ is the second-order moment for adaptive step sizing.
    \item $\exp_{\mathbf{x}_t}(\cdot)$ is the exponential map, which ensures the new $\mathbf{x}_{t+1}$ remains on $\mathcal{M}$.
\end{itemize}

\vfill
\pagebreak
\section{Experiments}
\label{app:experiments}
Whenever possible, we have tried to validate results as reported in the literature, as well as our theoretical contributions, to confirm that we have correctly implemented our algorithms. In this section, we cover:
\begin{enumerate}[noitemsep,topsep=0pt]
    \item Validating \path{Manify.embedders.delta_hyperbolicity} against a result in \citet{khrulkov};
    \item Validating \path{Manify.embedders.coordinate_learning} against a result in \citet{gu_learning_2018}; and
    \item Validating the relationships in Figure~\ref{fig:model_relationships} by directly contrasting various $\kappa$-GCN models with their Pytorch and \sklearn\ equivalents.
\end{enumerate}

\subsection{Validating $\delta$-Hyperbolicity Estimation (Khrulkov et al. (2020))}
We repeated the $\delta$-hyperbolicity measurements performed in \citet{khrulkov} on the ImageNet-10 and ImageNet-100 datasets \citep{deng2009imagenet} using all three encoders they used: VGG-19 \citep{simonyan2015very}, ResNet-34 \citep{he2016deep}, and Inception V3 \citep{szegedy2016rethinking}. For speed, we only report relative $\delta$-hyperbolicities ($\delta_\text{rel}$), which is defined as 
\begin{equation}
\delta_\text{rel} = \delta \frac{2}{\max_{i,j}{\D}},
\end{equation}
where $\D$ is the matrix of all pairwise distances.
Since we only report the maximum over a subset of the data, it is expected that our results slightly underestimate the true $\delta_\text{rel}$, as it is a maximum over a larger dataset.

\begin{table}[h]
    \centering
    \caption{Compaing $\delta$-hyperbolicity estimates for CIFAR-10 and CIFAR-100 test sets with \citet{khrulkov}. Our $\delta_\text{rel}$ is consistently (slightly) less than or equal to theirs, which is consistent with their $\delta_\text{rel}$ being a maximum taken over a larger dataset.}
    \begin{small}
    \begin{tabular}{cccc}
        \toprule
        Dataset & Encoder & Our $\delta_\text{rel}$ & \citet{khrulkov} $\delta_\text{rel}$\\
        \midrule
        CIFAR-10
            & VGG19 & 0.21 & 0.23\\
            & ResNet34 & 0.25 & 0.26\\
            & Inception v3 & 0.22 & 0.25\\
        \midrule
        CIFAR-100 
            & VGG19 & 0.22 & 0.22\\
            & ResNet34 & 0.27 & 0.25\\
            & Inception v3 & 0.22 & 0.23\\
        \bottomrule
    \end{tabular}
    \end{small}
    \label{tab:delta_hyperbolicity_validation}
\end{table}

\subsection{Validating Coordinate Learning (Gu et al 2018)}
\begin{figure}[h]
    \centering
    \includegraphics[width=\textwidth]{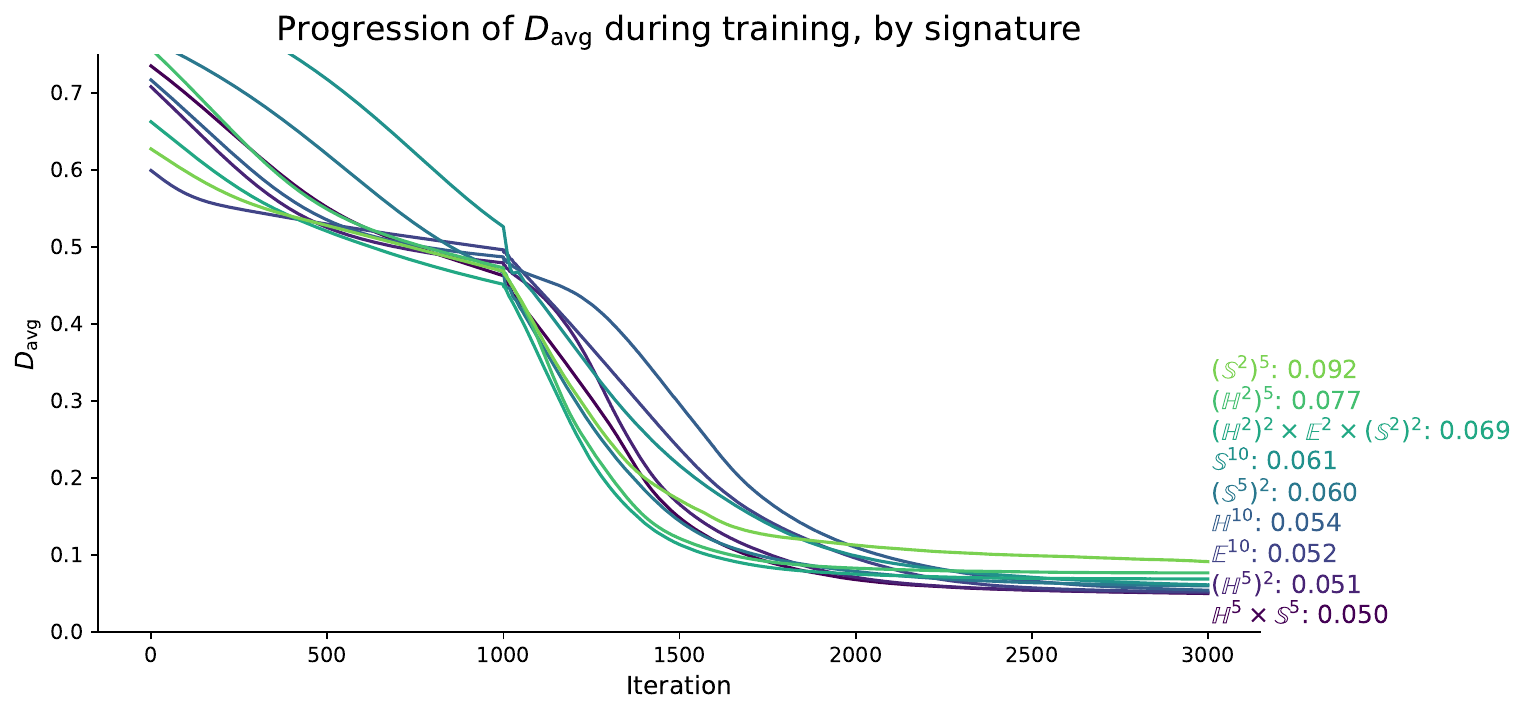}
    \caption{Loss curves for the embeddings learned in Table~\ref{tab:gu_reproduction}. Final $\Davg$ values are repeated in the legend on the right-hand side. Note that the loss curves suggest that all embeddings have converged.}
    \label{fig:davg_progression}
\end{figure}

To validate our implementation of the coordinate learning algorithm in \citet{gu_learning_2018}, we trained embeddings for the CS-PhDs dataset and compared our final $\Davg$ values to those reported in the paper.
We trained our model for 3,000 epochs, of which the first 1,000 were burn-in epochs, with a burn-in learning rate of 0.001, a training learning rate of 0.01, and learnable curvatures.
We show our final distortion values in Table~\ref{tab:gu_reproduction}, and plot the loss curve for each embedding in Figure~\ref{fig:davg_progression}.

\begin{table}[h]
    \centering
    \caption{Final $\Davg$ values for the CS-PhDs dataset as reproduced by us (left) and as reported in \citet{gu_learning_2018} (right).}
    \begin{small}
    \begin{tabular}{ccc}
        \toprule
        Signature & Our $\Davg$ & \citet{gu_learning_2018} $\Davg$\\
        \midrule
        $\E{10}$ & 0.0521 & 0.0543 \\
        $\H{10}$ & 0.0544 & 0.0502 \\
        $\S{10}$ & 0.0614 & 0.0569 \\
        $(\H{5})^2$ & 0.0509 & 0.0382 \\
        $(\S{5})^2$ & 0.0601 & 0.0579 \\
        $\H{5} \times \S{5}$ & 0.0501 & 0.0622 \\
        $(\H{2})^5$ & 0.0768 & 0.0687 \\
        $(\S{2})^5$ & 0.0915 & 0.0638 \\
        $(\H{2})^2 \times \E{2} \times (\S{2})^2$ & 0.0689 & 0.0765 \\
        \bottomrule
    \end{tabular}
    \end{small}
    \label{tab:gu_reproduction}
\end{table}

\vfill
\pagebreak
\subsection{Validating $\kappa$-GCN Model Relations}
\label{app:kappagcn_validation}
In Table~\ref{tab:kappagcn_validation}, we test the relationships laid out in Figure~\ref{fig:model_relationships}.
We applied the Wilcoxon rank-sum test over prediction accuracies for Gaussian mixtures with 3 classes in $\E{2}$ for 10 trials apiece.
We define the models as follows:
\begin{itemize}[noitemsep,topsep=0pt]
    \item The $\kappa$-GCN uses a radial basis function rescaled by 0.1 for $\hat{\mathbf{A}}$. This scaling factor was chosen to ensure $\hat{\mathbf{A}}$ does not resemble $\mathbf{I}$, preventing accidental resemblance to MLPs. We use four $\kappa$-GCN layers, followed by a stereographic logits layer;
    \item The $\kappa$-MLP is a $\kappa$-GCN with $\kappa=0$ and $\Ahat = \I$, and four hidden layers;
    \item The $\kappa$-MLR is a $\kappa$-GCN with $\kappa=0$, $\Ahat = \I$, and no hidden layers;
    \item The Pytorch GCN implements simple graph convolution layers with an analogous architecture and $\hat{\mathbf{A}}$ to $\kappa$-GCN. To match $\kappa$-GCN, it has square weights and no biases;
    \item The Pytorch MLP is a Pytorch neural network with four hidden layers. To match $\kappa$-GCN, it has square weights and no biases;
    \item The Pytorch MLR is a single linear layer,
    \item The ``\sklearn\ MLR'' refers to the `LogisticRegression' class in \sklearn with no regularization.
\end{itemize}
All predictors were trained for 1,000 epochs with a learning rate of 0.01 (Pytorch models) and 0.025 ($\kappa$-GCN derived models).
We use 0.025 to counteract the fact that $\kappa$-GCN rescales logits by a factor of 4, making the effective learning rates equivalent.

\begin{table}[h]
    \centering
    \caption{Wilcoxon rank-sum test $p$-values for accuracies of different models across 10 trials of Gaussian mixture classification. By default, we expect all GCN, MLP, and MLR models to be similar to one another (high $p$-value, and dissimilar to all other models. We color $p$-values less than 0.1 in red, and $p$-values greater than 0.9 in green. The observed green clusters reflect our expectations, except that \sklearn\ MLR diverges from $\kappa$- and Pytorch MLR.}
    \begin{small}
    \begin{tabular}{rccccccc}
        \hline
        & $\kappa$-GCN & Pytorch & $\kappa$-MLP & Pytorch & $\kappa$-MLR & Pytorch & \sklearn\\
        & & GCN & & MLP & & MLR & MLR \\
        \hline
        $\kappa$-GCN &   \cellcolor{green!25}1 &	\cellcolor{green!25}1 &	\cellcolor{red!25}0.0625 &	0.375 &	\cellcolor{red!25}0.0098 &	\cellcolor{red!25}0.0098 &	\cellcolor{red!25}0.0098\\
        Pytorch GCN &   \cellcolor{green!25}1 &	\cellcolor{green!25}1 &	0.7422 &	0.125 &	\cellcolor{red!25}0.0059 &	\cellcolor{red!25}0.0059 &	\cellcolor{red!25}0.0059\\
        $\kappa$-MLP &   \cellcolor{red!25}0.0625 &	0.7422 &	\cellcolor{green!25}1 &	\cellcolor{green!25}0.9453 &	\cellcolor{red!25}0.082 &	\cellcolor{red!25}0.082 &	\cellcolor{red!25}0.084\\
        Pytorch MLP &   0.375 &	0.125 &	\cellcolor{green!25}0.9453 &	\cellcolor{green!25}1 &	\cellcolor{red!25}0.0488 &	\cellcolor{red!25}0.0488 &	\cellcolor{red!25}0.0488\\
        $\kappa$-MLR &   \cellcolor{red!25}0.0098 &	\cellcolor{red!25}0.0059 &	\cellcolor{red!25}0.082 &	\cellcolor{red!25}0.0488 &	\cellcolor{green!25}1 &	\cellcolor{green!25}1 &	0.5547\\
        Pytorch MLR &   \cellcolor{red!25}0.0098 &	\cellcolor{red!25}0.0059 &	\cellcolor{red!25}0.082 &	\cellcolor{red!25}0.0488 &	\cellcolor{green!25}1 &	\cellcolor{green!25}1 &	0.625\\
        \sklearn\ MLR &   \cellcolor{red!25}0.0098 &	\cellcolor{red!25}0.0059 &	\cellcolor{red!25}0.084 &	\cellcolor{red!25}0.0488 &	0.5547 &	0.625 &	\cellcolor{green!25}1\\
        \hline
    \end{tabular}
    \label{tab:kappagcn_validation}
    \end{small}
\end{table}

\end{document}